%% file: for_arxiv.tex
\documentclass{navercloud}

\usepackage[T1]{fontenc}
\usepackage{textcomp}
\usepackage{lmodern}
\usepackage{pifont}
\usepackage{colortbl}
\usepackage{algorithm}
\usepackage{algpseudocode}
\usepackage{listings}
\usepackage{placeins} %
\usepackage{float} %
\usepackage{amsmath}

\graphicspath{{figures/}}
\input{content/macros.tex}

\renewcommand{\topfraction}{0.85}

\title{Evaluating Multimodal LLMs as Generalist Vision-Language-Action Agents for Drone Control: Commanding, Approaching, Tracking and Searching}

\author[\ast]{Jaewoo Park}
\author[\ast]{Minyoung Lee}
\author[\ast]{Sukmin Seo}
\author{Moonbin Yim}
\author{Hyunwook Yoon}
\author{Dohoon Ryu}
\author{Daehee Kim}
\author{Myungseo Song}
\author{Jihyuk Byun}
\author{Seunggyu Chang}
\author{Taeho Kil}
\author{Jiseob Kim}
\author{Bado Lee}
\author[\dagger]{Geewook Kim}

\affiliation{Drone AI Team, NAVER Cloud\\[2pt]
  {\small\texttt{\{jw0611.park, min.young.lee, sukmin.seo, gw.kim\}@navercorp.com}}\\[4pt]}

\contribution[\ast]{Equal contribution}
\contribution[\dagger]{Corresponding author}

\metadata[\optimisticfont GitHub]{\href{https://github.com/naver-ai/DroneCATS}{\texttt{github.com/naver-ai/DroneCATS}}}

\abstract{\input{content/abstract.tex}}

\begin{document}
\maketitle

\input{content/intro.tex}
\input{content/agent.tex}
\input{content/benchmark.tex}
\input{content/experiments.tex}

\input{content/conclusion.tex}

\section*{Code}
The code will be available at \url{https://github.com/naver-ai/DroneCATS}. Appendix~\ref{app:bench}
gives the maps, prompts and metric definitions needed to reproduce a run.

\bibliography{references}
\bibliographystyle{plainnat}

\appendix
\input{content/appendix.tex}

\end{document}

%% file: content/macros.tex
\newcommand{\bench}{DroneCATS}
\newcommand{\agent}{DroneCATS Agent}

\definecolor{cvGreen}{HTML}{059669}
\definecolor{cvAmber}{HTML}{D97706}
\definecolor{cvGray}{HTML}{94A3B8}
\newcommand{\yes}{\textcolor{cvGreen}{\ding{51}}}
\newcommand{\parti}{\textcolor{cvAmber}{(\ding{51})}}
\newcommand{\no}{\textcolor{cvGray}{\ding{55}}}

\lstdefinestyle{promptstyle}{
  basicstyle=\ttfamily\scriptsize,
  breaklines=true,
  breakatwhitespace=false,
  postbreak=\mbox{\textcolor{cvGray}{$\hookrightarrow$}\space},
  columns=fullflexible,
  keepspaces=true,
  showstringspaces=false,
  frame=single,
  rulecolor=\color{cvGray},
  framexleftmargin=3pt,
  framexrightmargin=3pt,
  xleftmargin=3pt,
  xrightmargin=3pt,
  aboveskip=1.2em,
  belowskip=0.6em,
  captionpos=b,
}

%% file: content/abstract.tex
Multimodal large language models (MLLMs) are increasingly used to make decisions for drones, but existing evaluations are mostly limited to simulation, to a single drone, and to targets that are visible from the start and do not move.
We present \bench{}, a benchmark in which the MLLM alone controls the drone through four actions, and evaluate ten MLLMs from frontier models to small open models.
The benchmark varies target visibility, target motion, scene and simulator, and extends to real drones.
Beyond a single drone, it adds a multi-drone suite, where one model drives multiple drones, and a commander + sub-agents suite, where a commander model directs one sub-agent for each of the drones.
Across these settings, we find that frontier models achieve the highest success rates, but their latency causes them to lose moving targets, while small open models often do not declare arrival even when they reach the target.
Guided by this failure analysis, we fine-tune a 2B model on drone action data synthesized from public RGB-D images; its success rate rises from 0\% to 31\% in simulation and outperforms the frontier models in real flight.
With multiple drones, we find that every zero-shot model succeeds more often as the commander of one sub-agent per drone than when it drives all of them from one context, and that commanding success follows general VLM ability.
On real drones, we observe that the results broadly follow those in simulation, while latency and hardware limits matter more.
We will fully open-source the benchmark.

%% file: content/intro.tex
\section{Introduction}
\label{sec:intro}

As Multimodal Large Language Models (MLLMs) have demonstrated strong performance on general vision-language tasks~\citep{liu2023llava,liu2023improvedllava,qwen2023qwenvl,wang2024qwen2vl,zhang2024llavvideo,bai2025qwen3vl,kim-seo-2024-efficient,kim2026mambamia}, they have been applied to a variety of domains.
One of these is to serve as vision-language-action (VLA) agents, in which the model receives a visual observation and a text instruction, selects an action, executes it, and receives the next observation.

In particular, MLLMs have recently been applied to drone control (Appendix~\ref{app:related}).
However, existing work has three limitations.
First, most existing benchmarks are evaluated only in simulation~\citep{liu2023aerialvln,xiao2025uavon,zhang2026esarbench,navaratnarajah2026zero}, so whether their results transfer to a real drone remains unknown.
Second, prior work largely focuses on a single drone~\citep{hu2025spf,zheng2026onfly,xu2026fly0,navaratnarajah2026zero}, although many real missions require several drones.
Third, single-drone studies typically assume that the target is visible from the start and remains stationary~\citep{navaratnarajah2026zero}, or that the route is given in the instruction~\citep{liu2023aerialvln,xu2026fly0}.

\begin{figure}[t]
\centering
\includegraphics[width=0.95\textwidth]{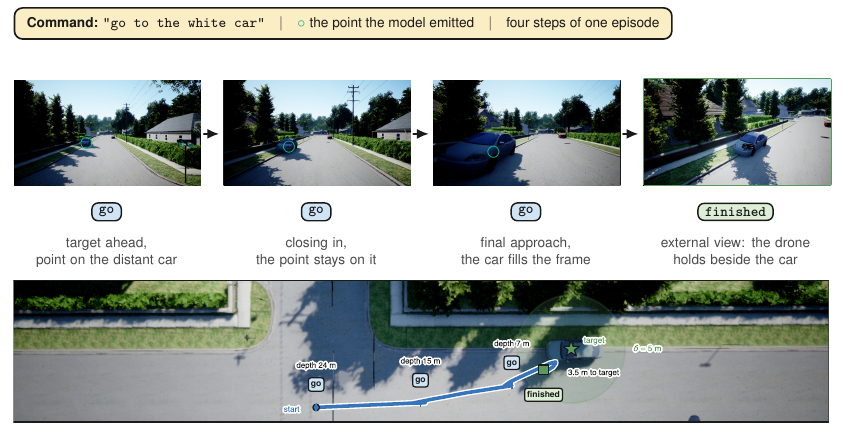}
\caption{An example of single-drone approach episode. The model receives the instruction and the
egocentric frame and returns one action per step; the circle marks the point it
emitted, and the fourth panel is an external view of the drone holding position
after it declared arrival. The band below shows the same flight from above, with
the $\delta$ ring on the target and the declaration marked.}
\label{fig:teaser}
\end{figure}

We therefore introduce \bench{}: a benchmark in which the MLLM is the sole decision maker of the drone.
Given a user's instruction (e.g., ``go to the white car''), the model decides at each step where to fly, when to look around, when to stop and think, and when to declare that it has arrived (Figure~\ref{fig:teaser}).
A model-agnostic controller executes the action, and the episode ends when the model declares that the instruction is complete.
We evaluate this ability under a wide range of conditions: target visibility and motion, diverse scenes across simulators, and real drones.

The same agent extends to multiple drones without modification.
We first design a multi-drone structure, in which one MLLM receives the frames of all drones and controls all of them in a single call.
We further design a commander + sub-agents structure, in which a commander MLLM writes commands for multiple sub-agents, and each sub-agent flies one drone.
In both structures, the instruction is one that no single drone can complete: the drones have to split up and gather what they find.

We evaluate $10$ MLLMs, from frontier API models to small open models.
The task is far from solved, and the strongest model completes about two thirds of the single-drone episodes.
The failures differ by model: the frontier models recognise the target but respond too slowly to follow a moving target, while the small open models reach the target but fail to follow the action protocol in the system prompt and therefore rarely declare that they have arrived.
Since these two failures limit MLLM pilots, we fine-tune the smallest and fastest 2B model on drone action data synthesised from public RGB-D images.
With the simulator and the real drone unseen in training, its success rate rises from 0\% to 31\% in simulation, and it outperforms the frontier models on a real drone.

With multiple drones, the command structure matters more than model size: every zero-shot model does better as a commander giving orders to sub-agents than as a pilot flying the drones itself.
The gap between commanders widens with difficulty: at the highest level, where the drones have to gather their readings and reason over them, only the models with the broadest general ability succeed.
Finally, the simulation results broadly carry over to real drones, where latency and hardware limits count for even more.

Our contributions are:
\begin{itemize}
\item \textbf{Agent.} \agent{} makes the MLLM the sole decision maker of a drone through $4$ prompt-defined actions. It covers out-of-view and moving targets, and scales to multi-drone and commander + sub-agents structures (Section~\ref{sec:agent}).
\item \textbf{Benchmark.} \bench{} scores the agent under one declaration-based criterion, and includes diverse scenes across multiple simulators (Section~\ref{sec:benchmark}).
\item \textbf{Evaluation.} $10$ MLLMs, from frontier API models to small open models and our fine-tuned 2B model, are evaluated in every suite (Section~\ref{sec:experiments}).
\item \textbf{Real-drone validation.} The single-drone and commander + sub-agents suites are flown on real drones, and the results broadly follow simulation (Section~\ref{sec:experiments}).
\end{itemize}

%% file: content/agent.tex
\section{\agent{}}
\label{sec:agent}

\subsection{Agent Loop and Action Space}
\label{sec:agent:action}

\begin{figure}[t]
\centering
\includegraphics[width=\textwidth]{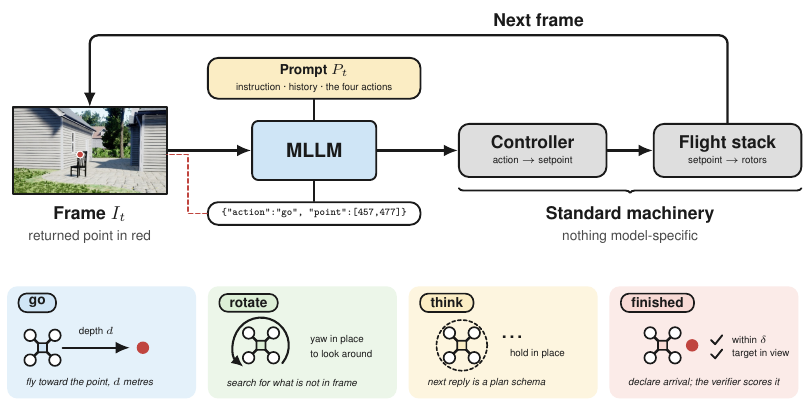}
\caption{The \agent{} loop (top) and its $4$ actions (bottom).}
\label{fig:system}
\end{figure}

Figure~\ref{fig:system} (top) shows one step of the control loop.
(1) The drone captures a frame $I_t$ from its forward camera.
(2) The prompt $P_t$ has $2$ parts: a system prompt, which defines the $4$ actions and is fixed in all episodes, and a user prompt, which includes the user's instruction (``go to the black chair'' in the figure), the current frame $I_t$ and the previous $5$ actions.
(3) The model returns one action as a JSON object.
(4) A rule-based controller converts the action into a setpoint, and the flight stack executes it.
(5) The next frame $I_{t+1}$ is captured from the new pose, and the action is appended to the history.
(6) The loop repeats until the model decides that it has reached the target and declares arrival.

Figure~\ref{fig:system} (bottom) shows the $4$ actions that the system prompt defines.
The definitions are written in natural language, and the model returns one of the $4$ as a JSON object (the full prompt and the execution rules in Appendix~\ref{app:bench}).

\textbf{\texttt{go}.}
The drone moves toward a point in the image.
The model returns the point $(x, y)$, normalised to $0$--$1000$, and a distance $d$ in metres; the point defines a camera ray, and the controller advances the drone $d$ along it, up to a step cap.
This is the only action that moves the drone.
Used when the target, or a path toward it, is visible.

\textbf{\texttt{rotate}.}
The drone turns in place.
The model returns a yaw angle $\theta$ within $\pm 90^\circ$.
Used to look around when the target is not in view.

\textbf{\texttt{think}.}
The drone holds position while the model deliberates.
On the next call the model fills a short plan schema (target in view, target match, closing in, current frame, action planning) before deciding on its action.
Used when the view is unclear, there are several candidates for the target, or the model is unsure what to do next.

\textbf{\texttt{finished}.}
The drone holds position and declares arrival.
The declaration is what the verifier scores.
Used when the model decides that the drone is close enough to the target.

\subsection{Task Axis}
\label{sec:agent:tasks}

\begin{figure}[t!]
\centering
\includegraphics[width=\textwidth]{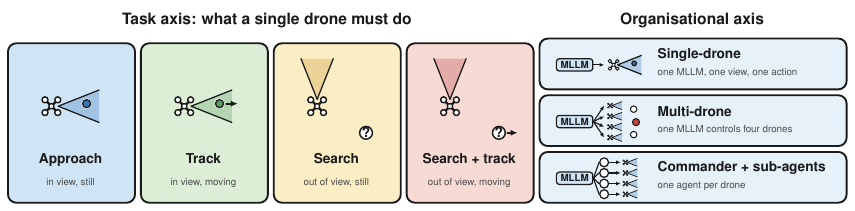}
\caption{The $2$ \bench{} axes. Left, the task axis (Section~\ref{sec:agent:tasks}): target motion and initial visibility give $4$ task types; the wedge is the drone's field of view, the filled dot a visible target, and the ring with a question mark a target whose location is unknown, so the drones of the two out-of-view tasks start facing away from it. Right, the organisational axis (Section~\ref{sec:agent:org}): a single drone, one MLLM commanding four drones from one context, and one agent per drone under a commander.}
\label{fig:taskgrid}
\end{figure}

\begin{figure}[t]
\centering
\includegraphics[width=0.48\textwidth]{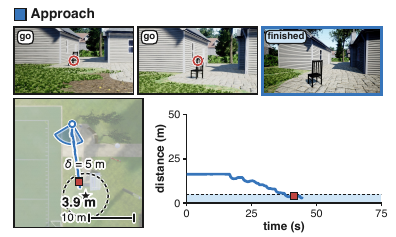}\hfill
\includegraphics[width=0.48\textwidth]{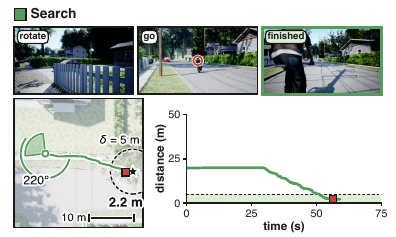}\\[6pt]
\includegraphics[width=0.48\textwidth]{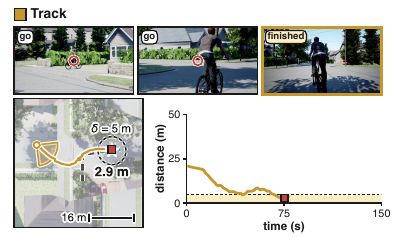}\hfill
\includegraphics[width=0.48\textwidth]{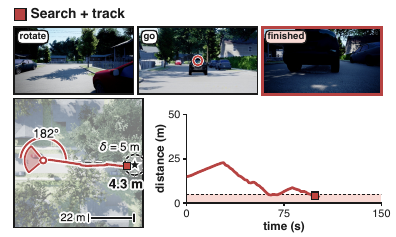}
\caption{One successful episode per task type, same map and same model.
Each panel shows $3$ egocentric frames, cropped toward the emitted point, above a top-down
trace on an aerial photograph. The wedge is the camera's field of view in the
first frame; its radius is symbolic. A thin arc marks the rotation before
departure where one occurred. The windows span $28.7$, $28.7$, $46$ and $62$\,m;
each panel carries a scale bar, and the $\delta = 5$\,m ring is drawn to scale.
The distance graphs end at 75\,s (top row) and 150\,s (bottom row).}
\label{fig:qual:tasks}
\end{figure}

All tasks require reaching a designated target and declaring arrival.
The task type is set by $2$ binary factors (Figure~\ref{fig:taskgrid}): whether the target is visible in the first frame, and whether it moves.
\emph{Approach} (visible, static): the target is in the first frame and does not move; this is the setting in which \citet{xu2026fly0} report backbone insensitivity.
\emph{Search} (out of view, static): the start pose is rotated away from the target, so the agent must sweep before approaching.
\emph{Track} (visible, moving): the target moves along a one-way route (speeds in Appendix~\ref{app:bench}).
\emph{Search + track} (out of view, moving): the target is initially out of view and moves; the agent must find it and then follow it.
Figure~\ref{fig:qual:tasks} shows one episode of each type.

\subsection{Organisational Axis}
\label{sec:agent:org}

\textbf{Single-drone.}
One MLLM controls one drone.
At each step it sees its own frame and returns one action; this is the loop of Section~\ref{sec:agent:action} unchanged.
The instruction names a single target, such as ``go to the black chair'', and the $4$ task types of Section~\ref{sec:agent:tasks} are evaluated in this setting.

\textbf{Multi-drone.}
One MLLM controls $N$ drones directly.
At each step, the $N$ frames of the $N$ drones are given in one prompt, and the model returns $N$ actions at once, one per drone.
One model call thus moves all $N$ drones by one step.

\textbf{Commander + sub-agents.}
There are $N{+}1$ MLLMs: one commander and $N$ sub-agents that each control one drone directly.
The commander sees the $N$ frames, each sub-agent's last action and a top-down map, and breaks the user's instruction down into one command per drone, such as ``go to the sign at the far left and read it''.
The commander repeats this every few steps and whenever a sub-agent returns \texttt{finished}.
A sub-agent is exactly the single-drone agent above: same prompt, same $4$ actions, its own frame only.
The one difference is that its instruction is the commander's command instead of the user's instruction.
Thus, the sub-agents fly the drones step by step, and the commander only decides where each drone goes.

The multi-drone and commander + sub-agents structures share their tasks, which come in $3$ difficulty levels for the commander: easy, medium and hard.
\begin{itemize}
\item \emph{Easy} --- License plate matching: one of $4$ cars carries the plate named in the instruction; a plate is readable only up close, so each drone approaches a car and reads it on its own, and the commander has little to do.
\item \emph{Medium} --- Sign matching: $4$ corridors each end in a sign, and the drones must reach the one that matches the instruction; the commander has to divide the corridors.
\item \emph{Hard} --- Three-sign sum: $3$ red signs show numbers, and the drones must read all $3$, add them, and enter the corridor whose entrance sign shows the sum, to the small sign at its end; the commander has to divide the search, gather the readings, and reason about the sum before sending the drones again.
\end{itemize}

%% file: content/benchmark.tex
\section{\bench{} Benchmark}
\label{sec:benchmark}

\bench{} evaluates the agent of Section~\ref{sec:agent} along its $2$ axes, one criterion for every episode.

\subsection{Metrics}
\label{sec:benchmark:verifier}

An episode succeeds if at least one arrival declaration was made while the drone was within a radius $\delta$ of the target in three dimensions and the target was visible to the camera:
\[
\text{success} \;=\; \exists\, i \;\in\; \text{declarations} \;:\; \mathrm{dist}_i \le \delta \;\wedge\; \mathrm{visible}_i .
\]
Every declaration is evaluated, and $\delta$ is set per episode (Appendix~\ref{app:bench}).
With $N$ drones, the criterion is applied to the pooled declarations: an episode succeeds when any drone declares arrival within $\delta$ of the target while the target is visible.
Judging is online: the declaration, the distance and the visibility are checked while the episode runs, and a success ends the episode early.
We report the success rate (SR) and the oracle success rate (OSR), the fraction of episodes whose trajectory entered $\delta$ at any step regardless of declaration; further metrics and the verdict in pseudocode are in Appendix~\ref{app:verifier}.

\subsection{Maps and Episodes}
\label{sec:benchmark:setup}

Episodes run in AirSim and Isaac Sim~\citep{shah2017airsim,gao2026nvidia} (Appendix~\ref{app:isaac}).
The maps cover indoor and outdoor scenes: a residential neighbourhood with house interiors, a campus, and the AirSim block map. The Isaac Sim suite uses one outdoor street map, Brownstone.
The single-drone suite has $20$ episodes per task type, $80$ in total, and the multi-drone and commander + sub-agents suites $3$ difficulty levels of $10$ episodes each at $N{=}4$, flown under every command structure and commander (Section~\ref{sec:experiments:org}).
Simulated episodes are capped at $300$\,s; the target placement and $\delta$ differ between outdoor, indoor and real-flight episodes (Appendix~\ref{app:bench}).

%% file: content/experiments.tex
\section{Experiments}
\label{sec:experiments}

\subsection{Single-Drone Results}
\label{sec:experiments:one}
\label{sec:experiments:main}
\label{sec:benchmark:models}

\textbf{Setup.}
Each episode flies one drone that must reach the designated target and declare arrival, on the $4$ task types of Section~\ref{sec:agent:tasks}, with $20$ episodes per cell, each flown twice.
Ten models are evaluated (Appendix~\ref{app:serving}): four frontier closed models, five open models, and one model fine-tuned by us.

\textbf{Success rates.}
\label{sec:experiments:sr}
The left side of Table~\ref{tab:main} reports SR per task type.
First, Gemini 3.7 Flash has the highest mean SR: it both reaches the target and declares arrival.
Second, the best model depends on the task type: Gemini 3.7 Flash on the two static-target tasks, Gemini Robotics-ER 2 on \emph{track} and GPT-5 on \emph{search + track}.
Since all four frontier models ground the target, the difference on moving targets comes from latency: GPT-5 and Gemini Robotics-ER 2 return an action in $1.4$ and $2.0$\,s per step and the other two in $2.9$ and $3.5$\,s (Table~\ref{tab:main}; latency includes serving, Appendix~\ref{app:serving}), and a faster model keeps up with a moving target.
Third, SR scales with model size: within the Qwen3.5 family, mean SR increases monotonically from 2B to 27B.

\begin{table}[t]
\centering
\caption{Single-drone results (\%; $2$ maps, $10$ episodes each, flown twice); frontier models, open models and our fine-tuned model separated by rules. Left, SR per task type over 20 episodes per cell, each flown twice. Right, over all $160$ runs of the model: (1) SR, the mean success rate. (2) OSR, the share of runs whose trajectory entered $\delta$ at any step, regardless of declaration. (3) NE, the median navigation error in metres, the distance to the target at the last action of the run. (4) Lat., the median model latency per step, in seconds. Best per column in bold.}
\label{tab:main}
\input{tables/table_main.tex}
\end{table}

\textbf{Recognition and declaration.}
\label{sec:experiments:gap}
We examined the failure cases, and they divide into two kinds: not recognising the target and not declaring arrival.
The right side of Table~\ref{tab:main} tells them apart.
The OSR--SR gap counts the episodes in which the model came within $\delta$ and did not declare \texttt{finished}.
NE, the distance at which an episode ends, shows how far from the target a model gets.
The open models of 9B and below turn under half of the episodes in which they reach $\delta$ into a success, so they fail at the declaration.
Among them, Qwen3.5-2B and zero-shot Cosmos3-Edge-2B also have the largest NE, $36.7$ and $77.7$\,m, farther than they started: when they do come close they do not declare, fly past the target and end far from it.
The frontier models and Qwen3.5-27B turn over three quarters into a success and have a small NE, hence a high SR.
\label{sec:experiments:taxonomy}
Other failure cases are in Appendix~\ref{app:qual}.

\textbf{System prompt following.}
\label{sec:experiments:spf}
We additionally evaluate the $10$ models on system-prompt-following and general VLM benchmarks.
Notably, SysBench~\citep{qin2024sysbench}, which asks a model to hold constraints given in a system prompt across a multi-turn conversation, correlates strongly with how much of a model's reaching becomes a declaration (Spearman $0.95$ over the nine off-the-shelf models), far more than with the reaching itself ($0.62$).
Detailed results are in Appendix~\ref{app:static}.

\textbf{Fine-tuning a drone agent.}
\label{sec:experiments:learn}
The observations above point to two advantages for a drone model: declaring arrival and deciding fast.
Cosmos3-Edge-2B has the lowest latency of the roster, $0.6$\,s per step (about $2$\,Hz on one L40S, Appendix~\ref{app:serving}), but never declares.
We therefore fine-tuned it on data we synthesise from public RGB-D datasets without any benchmark frame: a VLM tags each target and its pixel, the depth label is read from the ground-truth depth map, drone actions are derived by rule from the target's position and depth, and depth is expressed in a canonical camera so that the labels transfer across cameras (Appendix~\ref{app:bench:ft}).
The drone data are mixed with general instruction data.
Surprisingly, mean SR rises from $0$\% to $31.2$\%, above Qwen3.5-9B ($21.2\%$).
The OSR--SR gap also falls, from $26.2$ to $1.3$ points, so the model now declares when it reaches the target.
Its general-benchmark scores are preserved after fine-tuning; the full results are in Appendix~\ref{app:bench:ft}.

\input{content/multidrone}

\subsection{Real-Drone Flights}
\label{sec:experiments:real}

\textbf{Setup.}
We use a DJI Tello controlled through the Tello SDK, with the agent, prompt and action space of Section~\ref{sec:agent} (settings in Appendix~\ref{app:bench:real}).
The targets range from a toy plane, a laptop sleeve and an egg to an umbrella, a person and another drone.
The criterion of Section~\ref{sec:benchmark:verifier} is unchanged, with the distance and the visibility at each declaration judged in real time by a third-person observer; $\delta$ is $1.5$\,m, and the episode cap is $150$\,s for single-drone and $300$\,s for commander + sub-agents episodes.

\begin{figure}[t]
\centering
\begin{minipage}{\textwidth}{\sffamily\bfseries\small (a)}\hspace{3pt}{\sffamily\small Single drone, \emph{search + track}, Claude Opus 5}\\[2pt]
\includegraphics[width=\textwidth]{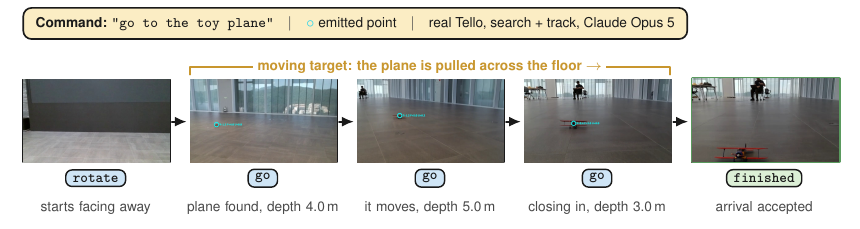}\end{minipage}\\[6pt]
\begin{minipage}{\textwidth}{\sffamily\bfseries\small (b)}\hspace{3pt}{\sffamily\small Two drones, Three-sign sum, Gemini 3.7 Flash commanding}\\[2pt]
\includegraphics[width=\textwidth]{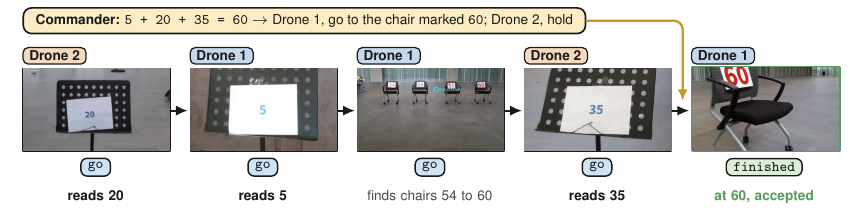}\end{minipage}
\caption{Real-drone episodes. (a) Single drone, \emph{search + track}: Claude Opus 5 starts facing away from a toy plane pulled across the floor, turns until it sees it, follows it and declares arrival once. (b) Two drones, Three-sign sum: Gemini 3.7 Flash commands two Tellos. Drone 1 reads 5 and finds the chairs, Drone 2 reads 20 and 35, and the commander sends Drone 1 to chair 60.}
\label{fig:real}
\end{figure}

\textbf{Single drone.}
We flew $5$ models: Gemini 3.7 Flash, Claude Opus 5, Qwen3.5-9B, and Cosmos3-Edge-2B zero-shot and fine-tuned.
Table~\ref{tab:real}(a) reports $100$ episodes, $5$ per cell.
Real and simulated success rates are correlated across the $20$ model-by-task cells (Spearman $\rho = 0.57$, $p < 0.01$).
As in simulation, Gemini 3.7 Flash is the best model on \emph{approach} and \emph{search}, zero-shot Cosmos3-Edge-2B never declares, and the fine-tuned Cosmos3-Edge-2B declares and reaches a mean SR of $55$\%.
Latency, however, matters more in real flight than in simulation.
The video and command links add a delay to every step.
Gemini 3.7 Flash and Claude Opus 5 track well in simulation but lose the moving target in real flight, so their \emph{track} and \emph{search + track} SRs are low.
Cosmos3-Edge-2B (FT) has a model latency of $0.6$\,s, and this advantage grows in real flight: it is the only model whose real mean SR exceeds its simulated one, $55$\% against $31.2$\%.
Figure~\ref{fig:real}(a) shows an example.

\textbf{Commander + sub-agents.}
We flew $3$ commanders: Gemini 3.7 Flash, Claude Opus 5 and Qwen3.5-9B.
The sub-agents are $2$ Tellos, each run by Gemini Robotics-ER~2 as in Section~\ref{sec:experiments:org}.
Table~\ref{tab:real}(b) reports $18$ episodes, $6$ per commander.
The task is the \emph{hard} level, Three-sign sum, without the map.
As in simulation, with or without the map, both frontier commanders are far ahead of Qwen3.5-9B.
Gemini 3.7 Flash has the highest SR, $83$\%, then Claude Opus 5, $50$\%, and Qwen3.5-9B, $17$\%.
Gemini 3.7 Flash also arrives faster than Claude Opus 5 in both real flight and simulation.
Of the $9$ failed episodes, $5$ did not find and read all $3$ numbers, $3$ misread or mis-added, and $1$ knew the answer but did not arrive.
We fly only $2$ drones in real flight, not the $4$ of simulation, because each Tello is a Wi-Fi access point with a fixed IP, and more than $2$ drones interfered.
Such hardware limits matter as much as the model. Figure~\ref{fig:real}(b) shows one flight.

\begin{table}[t]
\centering
\caption{Real-drone flights. (a) Single-drone: SR (\%) over $5$ episodes per cell, the row mean, and the mean SR of the same model in simulation. (b) Commander + sub-agents with $2$ Tello sub-agents on the \emph{hard} level, Three-sign sum: SR (\%) over $6$ episodes and the median time of the successful episodes, next to the same commanders in simulation with the map ($10$ episodes). Best SR per column in bold.}
\label{tab:real}
\begin{minipage}[t]{0.585\textwidth}\centering
\input{tables/table_real.tex}
\end{minipage}\hfill
\begin{minipage}[t]{0.395\textwidth}\centering
{\footnotesize\textit{(b) Commander + sub-agents}}\\[2pt]
\input{tables/table_multi_real.tex}
\end{minipage}
\end{table}

%% file: tables/table_main.tex
\newlength{\numw}
\setlength{\numw}{2.45em}
{\footnotesize
\setlength{\tabcolsep}{2pt}
\renewcommand{\arraystretch}{1.15}
\begin{tabular}{@{}l@{\hspace{5pt}}>{\centering\arraybackslash}p{5.8em}>{\centering\arraybackslash}p{4.2em}>{\centering\arraybackslash}p{3.8em}>{\centering\arraybackslash}p{4.6em}@{\hspace{6pt}}>{\centering\arraybackslash}p{2.4em}>{\centering\arraybackslash}p{2.4em}>{\centering\arraybackslash}p{2.4em}>{\centering\arraybackslash}p{2.4em}@{}}
\toprule
& \multicolumn{4}{c}{SR per task type} & \multicolumn{4}{c}{All 160 runs} \\
\cmidrule(lr){2-5}\cmidrule(lr){6-9}
Model                 & Approach & Search & Track & Search + Track & SR & OSR & NE & Lat. \\
\midrule
Gemini 3.7 Flash      & \textbf{87.5} & \textbf{67.5} & 55 & 47.5 & \textbf{64.4} & \textbf{72.5} & 4.3 & 3.5 \\
Claude Opus 5         & 57.5 & 47.5 & 40 & 40 & 46.2 & 51.2 & 5.2 & 2.9 \\
Gemini Robotics-ER 2  & 62.5 & 47.5 & \textbf{70} & 50 & 57.5 & 65 & 4.3 & 2.0 \\
GPT-5                 & 55 & 37.5 & 62.5 & \textbf{65} & 55 & 71.2 & \textbf{4.0} & 1.4 \\
\midrule
Qwen3.5-27B           & 60 & 45 & 25 & 27.5 & 39.4 & 40.6 & 7.2 & 1.0 \\
Qwen3.5-9B            & 32.5 & 15 & 32.5 & 5 & 21.2 & 63.1 & 13.7 & 1.4 \\
Qwen3.5-4B            & 12.5 & 5 & 7.5 & 0 & 6.2 & 18.8 & 16.7 & 1.0 \\
Qwen3.5-2B            & 0 & 0 & 0 & 0 & 0 & 13.8 & 36.7 & 0.8 \\
Cosmos3-Edge-2B       & 0 & 0 & 0 & 0 & 0 & 26.2 & 77.7 & \textbf{0.6} \\
\midrule
Cosmos3-Edge-2B (FT)  & 35 & 35 & 27.5 & 27.5 & 31.2 & 32.5 & 9.8 & \textbf{0.6} \\
\bottomrule
\end{tabular}}

%% file: content/multidrone.tex
\subsection{Multi-Drone and Commander + Sub-Agents Results}
\label{sec:experiments:org}

\textbf{Setup.}
Each episode flies four drones.
Table~\ref{tab:multi:pairs} covers $600$ episodes: $30$ per model, $10$ at each difficulty level, under either the multi-drone or the commander + sub-agents structure.
The commander is one of the $10$ models of Section~\ref{sec:experiments:one}, and the sub-agents are Gemini Robotics-ER~2 throughout.

\textbf{Commander + sub-agents beats multi-drone.}
On the same episodes, every zero-shot model succeeds more often as the commander of four sub-agents than when it drives the four drones itself ($\Delta$ in Table~\ref{tab:multi:pairs}).
The gain holds from the strongest model, Gemini 3.7 Flash ($\Delta = +43.3$ points), to the weakest, zero-shot Cosmos3-Edge-2B ($\Delta = +33.3$ points).
The reason differs by model size.
The frontier models have long latencies, and in multi-drone one inference has to produce all four actions, so one fleet step takes about $7$\,s and the fleet moves only about $8$ times per minute.
On the other hand, in commander + sub-agents, each sub-agent is Gemini Robotics-ER~2 and returns actions quickly, while the commander only re-plans, so the fleet moves $29$ times per minute within the same cap.
The small models fail for a different reason: they often return the same action for all four frames.
Qwen3.5-27B and Qwen3.5-9B return the same pixel and depth for all four drones in $75$\% and $43$\% of \texttt{go} actions, whereas the frontier models show no such pattern ($0$--$5$\%).
As a commander the same model only writes a text command per drone, and each sub-agent grounds the command in its own frame, so the same-action failure does not arise.

\begin{figure}[t]
\centering
\includegraphics[width=\textwidth]{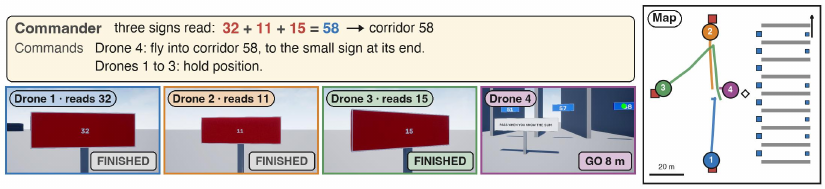}
\caption{The commander's view at one call on Three-sign sum. Upper left, what the commander concluded from the four views and the map, and the commands it sent. Lower left, each drone's view with the action its sub-agent took next. Drones 1 and 2 declared \texttt{finished} earlier, shown in grey, and Drone 3 declares at this call, so the commander tells Drones 1 to 3 to hold position. Right, the scene map with the four drones and their paths so far. Three drones are parked at the red signs they read. The commander adds the numbers and sends only Drone 4 into the corridor marked 58.}
\label{fig:multi:turn:main}
\end{figure}

\textbf{Differences across difficulty levels.}
The gap between commanders widens as the difficulty rises.
At the \emph{easy} level nearly every commander succeeds in $70$\% or more, and so does multi-drone: when the sub-agents can finish the task on their own, the commander does not matter.
At the \emph{medium} level the frontier models stay at $70$\% or more while the small models drop to $30$ to $80$\%: the small models fall behind once the commander has to spread the four drones over four corridors.
At the \emph{hard} level the gap is widest: Gemini 3.7 Flash and Claude Opus 5 succeed in $80$\%, and no other commander in more than $40$\%.
In all $16$ successes of Gemini 3.7 Flash and Claude Opus 5 the correct sum appeared in the commander's commands before the arrival (Figure~\ref{fig:multi:turn:main}), whereas the other commanders often arrived by chance, without a correct sum in their commands.
The control-tower map helps in the same place: across three commanders, SR on Three-sign sum rises from $23$\% to $57$\% with the map, while the two easier levels change little (Appendix~\ref{app:multidrone}).
The three levels show that the best commanders are those that can spread the drones and then gather what they read and reason over it.

\textbf{General knowledge as a commander.}
A commander's job is to work out the situation from the four views and turn the instruction into a command for each drone, which should relate to general ability.
We measured the $10$ models on a general VLM benchmark, MME~\citep{fu2026mme}, and on a multimodal instruction-following benchmark, MM-IFEval~\citep{ding2025mmifengine}.
Both scores correlate strongly with commander SR (Spearman $0.91$ and $0.95$ over the nine zero-shot commanders).
The frontier closed models, the best commanders, score high on both benchmarks, and the small open models, the weakest, score low.
Cosmos3-Edge-2B (FT), fine-tuned on drone action data, improves as a single-drone pilot but scores $3.3$\% as a commander: the data taught it to fly, not to command.

\begin{table}[t]
\centering
\caption{The same model in two roles: driving the four drones itself (multi-drone) and commanding four Gemini Robotics-ER~2 sub-agents (commander + sub-agents). Each difficulty level has ten episodes (easy: License plate matching, medium: Sign matching, hard: Three-sign sum), so a total is over $30$ episodes. Cells are SR (\%), and $\Delta$ is the gain in total SR points. Best per column in bold.}
\label{tab:multi:pairs}
\input{tables/table_multi_pairs.tex}
\end{table}

%% file: tables/table_multi_pairs.tex
\setlength{\tabcolsep}{4.5pt}
\renewcommand{\arraystretch}{1.12}
\footnotesize
\begin{tabular}{@{}lccccc>{\raggedleft\arraybackslash}p{2.4em}@{}}
\toprule
SR (\%) & Multi-drone & \multicolumn{4}{c}{Commander + sub-agents} & \\
\cmidrule(lr){2-2}\cmidrule(lr){3-6}
Model & Total & Easy & Medium & Hard & Total & $\Delta$ \\
\midrule
Gemini 3.7 Flash      & 50.0 & \textbf{100} & \textbf{100} & \textbf{80} & \textbf{93.3} & +43.3 \\
Claude Opus 5         & 46.7 & 90 & 90 & \textbf{80} & 86.7 & +40.0 \\
Gemini Robotics-ER 2  & \textbf{53.3} & \textbf{100} & 90 & 40 & 76.7 & +23.3 \\
GPT-5                 & 13.3 & \textbf{100} & 70 & 10 & 60.0 & \textbf{+46.7} \\
\midrule
Qwen3.5-27B           & 26.7 & 80 & 80 & 0 & 53.3 & +26.7 \\
Qwen3.5-9B            & 36.7 & 90 & 30 & 10 & 43.3 & +6.7 \\
Qwen3.5-4B            & 16.7 & 70 & 60 & 0 & 43.3 & +26.7 \\
Qwen3.5-2B            & 3.3 & 90 & 50 & 10 & 50.0 & \textbf{+46.7} \\
Cosmos3-Edge-2B       & 0.0 & 50 & 40 & 10 & 33.3 & +33.3 \\
\midrule
Cosmos3-Edge-2B (FT)  & 3.3 & 10 & 0 & 0 & 3.3 & 0.0 \\
\bottomrule
\end{tabular}

%% file: tables/table_real.tex
{\footnotesize\textit{(a) Single-drone}}\\[2pt]
{\scriptsize
\setlength{\tabcolsep}{2.4pt}
\renewcommand{\arraystretch}{1.05}
\begin{tabular}{@{}lcccccc@{}}
\toprule
& \multicolumn{5}{c}{Real} & Sim \\
\cmidrule(lr){2-6}\cmidrule(lr){7-7}
Model & Approach & Search & Track & \shortstack{Search\\+ Track} & Mean & Mean \\
\midrule
Gemini 3.7 Flash     & \textbf{100} & \textbf{80} & 0 & 0 & 45 & \textbf{64.4} \\
Claude Opus 5        & 40 & 40 & 20 & 20 & 30 & 46.2 \\
\midrule
Qwen3.5-9B           & 40 & 0 & 0 & 0 & 10 & 21.2 \\
Cosmos3-Edge-2B      & 0 & 0 & 0 & 0 & 0 & 0 \\
\midrule
Cosmos3-Edge-2B (FT) & 60 & 60 & \textbf{60} & \textbf{40} & \textbf{55} & 31.2 \\
\bottomrule
\end{tabular}}

%% file: tables/table_multi_real.tex
\setlength{\tabcolsep}{2.4pt}
\renewcommand{\arraystretch}{1.12}
\scriptsize
\begin{tabular}{@{}lcccc@{}}
\toprule
& \multicolumn{2}{c}{Real, 2 drones} & \multicolumn{2}{c}{Sim, 4 drones} \\
\cmidrule(lr){2-3}\cmidrule(lr){4-5}
Commander & SR & Time (s) & SR & Time (s) \\
\midrule
Gemini 3.7 Flash & \textbf{83} & 118 & \textbf{80} & 213 \\
Claude Opus 5    & 50 & 151 & \textbf{80} & 242 \\
Qwen3.5-9B       & 17 & 95 & 10 & 114 \\
\bottomrule
\end{tabular}

%% file: content/conclusion.tex
\section{Conclusion}
\label{sec:conclusion}
We presented \bench{}, a benchmark that evaluates MLLMs as the sole decision maker of a drone.
It covers approaching, searching and tracking with one drone and commanding several drones, in simulation and on real drones, all scored with one declaration-based criterion.
The results show that an MLLM can already control a drone from the prompt alone, and that its reliability depends not only on perception but also on declaring arrival and acting quickly.
Fine-tuning a small, fast model on synthesized data teaches it to declare arrival, which favours small on-device pilots over frontier APIs for real flight.
For several drones, structure matters more than the model: a commander writing text for per-drone agents lets every model do better, and commanding draws on the general ability that existing VLM benchmarks measure.
Because the simulation results carry over to real flight, \bench{} can serve as a proxy before flying.
Future work extends the benchmark from reaching a target to more general drone scenarios.

%% file: content/appendix.tex
\AddToHook{env/table/begin}{\setlength{\belowcaptionskip}{6pt}}
\renewcommand{\topfraction}{0.9}\renewcommand{\bottomfraction}{0.8}\renewcommand{\textfraction}{0.07}
\renewcommand{\floatpagefraction}{0.85}\setcounter{topnumber}{3}\setcounter{bottomnumber}{2}\setcounter{totalnumber}{5}

\section{Benchmark Details}
\label{app:bench}

\subsection{Maps and Episodes}

The AirSim \emph{single-drone} suite uses two maps, a residential neighbourhood and a campus, and each map contributes ten episodes per task cell, for twenty per cell and eighty in all. Every episode is flown twice and is capped at $300$\,s of simulated time, indoors and outdoors alike. The Isaac Sim suite is described in Appendix~\ref{app:isaac}.

The two maps name their targets differently.
On the residential map every target is an asset spawned into the scene, one per episode: vehicles, a cyclist, drones, a skateboard, furniture and small objects, named by type and, where needed, by colour (``go to the black SUV'').
On the campus map the static targets are seven fixtures native to the scene and three spawned drones. The fixtures are generic and often near-duplicates, so their referring expressions disambiguate rather than name a colour (``go to the bench beside the trash can on the west side of the field'').

The start pose and the target motion set the task type.
Search and search + track episodes reuse the start poses of approach and track with the yaw turned by $180^\circ$, so the target is behind the drone at the first frame.
Moving targets travel at $0.3$\,m/s outdoors and $0.15$\,m/s indoors along one-way routes.
Three of the ten static episodes and two of the ten moving ones on the residential map are set inside a house, where $\delta$ is $2.5$\,m instead of $5$\,m.

\begin{figure}[!htb]
\centering
\includegraphics[width=\textwidth]{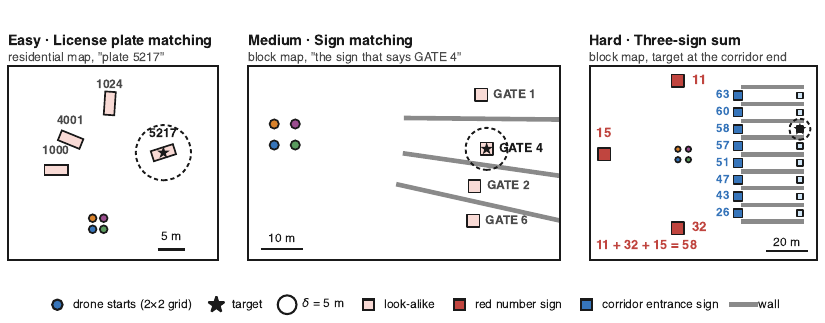}
\caption{One episode per category with four drones, seen from above and drawn from the episode records. The four drones start on a $2\times2$ grid, and the dashed ring is $\delta = 5$\,m around the target.}
\label{fig:multi:layout}
\end{figure}

The multi-drone and commander + sub-agents suites have ten episodes per category (Figure~\ref{fig:multi:layout}).
License plate matching uses the residential map, and Sign matching and Three-sign sum use the block map that ships with AirSim.
In License plate matching and Sign matching the target has three look-alikes that differ from it only in the inscription.
In Three-sign sum the fleet reads three red numbers and flies to the end of the corridor whose entrance sign shows their sum.
The verdict pools the declarations of the four drones (Section~\ref{sec:agent:org}).

\subsection{Execution}
\label{app:bench:exec}

The model outputs a point and a depth, and a rule-based controller turns them into flight. This section lists the constants of that controller.
The controller answers \texttt{go} with a turn and a move toward the point, \texttt{rotate} with a yaw in place, and \texttt{think} and \texttt{finished} by holding position.

\paragraph{Back-projection.}
\texttt{go} carries a point and a depth, and the controller back-projects the pair through a pinhole camera.
With image width $W$, height $H$ and horizontal field of view $\phi$, the focal length in pixels is $f = W / (2\tan(\phi/2))$.
The same value is used vertically, assuming square pixels, and the principal point is the image centre $(W/2, H/2)$.
A model point $(u, v)$ in $0$--$1000$ becomes the pixel $(p_x, p_y) = (uW/1000,\; vH/1000)$ and then the body-frame displacement
\begin{equation*}
d_x = \frac{p_x - W/2}{f}\, d, \qquad
d_y = d, \qquad
d_z = \frac{H/2 - p_y}{f}\, d,
\end{equation*}
with $d_x$ right, $d_y$ forward and $d_z$ up.
The depth $d$ is the distance along the optical axis, the convention of the simulator's \texttt{DepthPlanar} buffer, so the oracle depth logged at each step is directly comparable to the model's estimate.
The field of view is $\phi = 90^\circ$. The camera renders at $1280 \times 720$, and the model receives the frame resized to $640 \times 360$.

\paragraph{Executing \texttt{go}.}
For \texttt{go}, the drone first yaws in place if the horizontal bearing to the point exceeds an $8^\circ$ deadband, and it holds its altitude if the vertical component is less than $0.15$ of the horizontal distance.
It then flies toward the point with a body-frame velocity command (\texttt{moveByVelocityBodyFrameAsync}) held for at most $2.5$\,s at up to $1.2$\,m/s, so one step moves at most $3$\,m.
Two depth bands shrink the step near the target. Under $5$\,m the step is scaled by $0.6$, and under $2$\,m it is scaled to zero and the drone brakes, so a model that reports a very small depth stops instead of creeping forward.
These are the outdoor values. Indoor episodes cap the step at $1.5$\,m (at most $0.6$\,m/s), put the bands at $2.5$\,m and $1.0$\,m, and scale the airframe by $0.5$.

\paragraph{Executing \texttt{rotate}.}
For \texttt{rotate}, the drone yaws at $60^\circ$/s for $|\theta| / 60$ seconds, a rate command (\texttt{rotateByYawRateAsync}) rather than a goal-angle command.
The prompt asks for at most $90^\circ$ per step, and the parser clamps larger requests to $\pm 90^\circ$.

\paragraph{Simulator.}
AirSim~$1.8.1$ with its built-in \texttt{SimpleFlight} flight controller, not PX4 SITL, so no autopilot firmware sits between the controller and the airframe.
The forward camera is mounted $0.5$\,m ahead of and $0.1$\,m below the body origin with zero pitch.

\subsection{Serving and Decoding} \label{app:serving}

The open models range from a mid-size 27B model down to 2B models small enough to fly.
Serving Cosmos3-Edge-2B with a 32k context takes $10.5$\,GiB and $0.44$\,s per 30-token decode, about $2$\,Hz, measured with vLLM at bfloat16 on one L40S, and a model of this size fits within the memory of a Jetson Orin NX. Qwen3.5-4B needs $13.9$\,GiB even
though its hybrid linear attention makes context nearly free.
The 2B and 4B sizes were served locally, everything larger over an API (Table~\ref{tab:models}).

Every model ran through the same agent with an $8192$-token output cap and temperature $0.4$.
GPT-5 accepts only its default temperature, so none is sent to it.
On the Gemini path top-p is $0.95$ and top-k is $40$.
In the single-drone runs extended reasoning was set to the lowest level each API allows, and a \texttt{think} action switches the next call to a plan schema (Appendix~\ref{app:results}).
The multi-drone runs keep the same lowest level.
The commanders of Section~\ref{sec:experiments:org} run at the highest reasoning level at which one call returns within the lifetime of an order, six sub-agent steps (Appendix~\ref{app:multidrone}): a dynamic budget for Gemini 3.7 Flash, low effort for Claude Opus 5 and GPT-5, whose medium calls take $14$ to $21$\,s and outlast the orders, and off for the Qwen models, whose calls take $70$ to $240$\,s with reasoning on, and for Cosmos3-Edge-2B and Gemini Robotics-ER~2.

\begin{table}[!htb]
\centering
\caption{Evaluated models, grouped into frontier closed models, open models and our fine-tuned model. The identifier is the exact API or Hugging Face string, and the local models are served with vLLM 0.25.1. Results tables refer back to this one rather than repeating the citations.}
\label{tab:models}
\input{tables/table_models.tex}
\end{table}

\subsection{Real-Drone Flights}
\label{app:bench:real}
The platform is a DJI Tello (Boost Combo) with its $960\times720$ forward camera ($56.4^\circ$ horizontal field of view, measured by checkerboard calibration).
The agent, prompt, parser and episode runner are those of the simulation, and only the controller changes.
\texttt{go} is executed through the SDK \texttt{go} command at $60$\,cm/s for one step of at most $1.5$\,m. Within an estimated $2.5$\,m of the target the step is scaled to $0.6$, and within $1.0$\,m no forward motion is commanded.
\texttt{rotate} is executed in place, and \texttt{think} and \texttt{finished} hold position.
The drone takes off to $1.5$\,m and holds position after a \texttt{finished}. An observer then judges the declaration (within $\delta{=}1.5$\,m of the target and target visible in the frame) and either ends the episode or lets it continue, so several declarations can be scored per episode as in simulation.
Episodes are capped at $150$\,s of flight time, excluding time held for judging.
Each model flies five episodes per task type, one per target, in two indoor rooms.
Static targets (approach, search) are a gray water bottle, a pink laptop sleeve, a military uniform, a white umbrella and an egg, placed $5$--$10$\,m from the start pose.
Moving targets (track, search + track) are a person wearing the military uniform, a person carrying the pink laptop sleeve, a toy plane, a snack box and a second drone, moved during the episode.
The drone starts facing the target in approach and track and facing away from it ($180^\circ$) in search and search + track.
Every moving target follows a fixed route of $22.5$\,m over the $150$\,s cap, that is $0.15$\,m/s, the speed of the indoor moving targets in simulation.
A target is either pulled along the route on a line or carried by a person walking at that speed.

The \emph{commander + sub-agents} flights use two Tellos, each run by a Gemini Robotics-ER~2 sub-agent under one commander, with a $300$\,s cap.
The category is Three-sign sum without the map.
The three numbers are printed small in blue on sheets on three music stands, so a drone has to fly up to a stand to read one, and the four candidate answers are printed large in red on four chairs, so a drone that knows the sum can head for the right chair from across the room.
The small numbers test whether the fleet searches and reads, and the large ones keep a correct sum from failing on a second search.
Each of the three commanders flies six sets, one episode each, where every set is a different arrangement of the stands and chairs with different numbers, and the declarations are judged as in the \emph{single-drone} flights.

\subsection{The Prompt and Request Format}
\label{app:request}

The prompt $P_t$ of Section~\ref{sec:agent:action} has two parts.
The system part $P^{\mathrm{sys}}$ is Listing~\ref{lst:prompt} with the model's coordinate convention substituted, and it is sent as the system message.
The user part $P^{\mathrm{user}}_t$, which holds the \texttt{Command} line, the history and the closing line, is sent as the user message and is followed by the current frame $I_t$.
The frame keeps its aspect ratio, is resized only when it is wider than $640$ pixels, and is JPEG-encoded at quality $85$.

Three fields are substituted at every step.
\texttt{\{instruction\}} is the episode's referring expression, for example \emph{go to the black SUV}.
\texttt{\{point\_fmt\}} is \texttt{[x, y]} or \texttt{[y, x]} according to the coordinate convention the model was trained with, and the parser follows the same setting, so a model is never penalised for its own axis order.
\texttt{\{history\}} is the last five actions, oldest first, one per line, with the most recent tagged, written in the compact form the logs use.
Before the first step the field reads \texttt{(none yet --- this is the first step)}.
The history lets a model notice that it is circling, and nothing in the scaffold detects circling on its behalf.

Every request also carries the JSON schema $S_t$ that the reply must satisfy.
On an ordinary step this is the action schema (\texttt{action}, \texttt{point}, \texttt{depth\_m}, \texttt{angle\_deg}), and on the step after a \texttt{think} it is the plan schema of Listing~\ref{lst:plan-schema}.
OpenAI-compatible endpoints, including OpenAI, OpenRouter and the locally served models, receive the schema as \texttt{response\_format} (Listing~\ref{lst:openai-request}), and the Gemini endpoint as \texttt{response\_schema} (Listing~\ref{lst:gemini-request}).
No tools or function definitions are passed, so the schema is the only structured-output mechanism.
Formatting therefore separates no model, and the declaration failures in Section~\ref{sec:experiments:gap} are failures of the prompt, not of formatting.

\paragraph{How \texttt{think} is wired.}
Extended reasoning is kept at the lowest level each API allows (Appendix~\ref{app:serving}).
When the previous action was \texttt{think}, the current call answers the plan schema instead of the plain action schema.
The plan schema has five fields, in the order the model is to settle them, followed by the same action fields (Listing~\ref{lst:plan-schema}).
The three categorical fields are closed enumerations, and the two free-text fields are capped at $1{,}200$ characters.
Deliberation is thus carried as schema fields rather than native reasoning tokens, identically across providers.
We chose this over native reasoning because native reasoning is slow and some models expose no control over its effort.
On moving targets a single long reasoning call can be enough to lose the target, which drives the success rate down.
The schema keeps the length of the deliberation under the user's control.
The plan schema is on in all runs, and the system prompt then also carries a short rule describing the plan fields (\texttt{prompts/prompt\_system\_yx\_plan.txt} in the code release).

\input{content/prompt.tex}

\begin{lstlisting}[style=promptstyle, float=tp, caption={The plan schema, as sent after a \texttt{think}. \texttt{action} is required, and \texttt{point}, \texttt{depth\_m} and \texttt{angle\_deg} are the action's own arguments, as in the plain action schema.}, label={lst:plan-schema}]
{
  "target_in_view":  "yes" | "no" | "unsure",
  "target_match":    "match" | "mismatch" | "unsure",
  "closing_in":      "yes" | "no" | "unknown",
  "current_frame":   <string, 20-1200 chars>,
  "action_planning": <string, 20-1200 chars>,
  "action":   "go" | "rotate" | "think" | "finished",
  "point":    [x, y],        // 0-1000 image coordinates, go only
  "depth_m":  <0-200>,       // go only
  "angle_deg": <-90..90>     // rotate only
}
\end{lstlisting}

\begin{lstlisting}[style=promptstyle, float=tp, caption={Abbreviated OpenAI-compatible multimodal request. Provider-specific decoding fields are omitted.}, label={lst:openai-request}]
{
  "model": "<model identifier>",
  "messages": [
   {"role": "system", "content": "<P_sys>"},
   {"role": "user",
    "content": [
      {"type": "text", "text": "<P_user_t>"},
      {"type": "image_url",
       "image_url": {"url": "data:image/jpeg;base64,<base64(JPEG(I_t))>"}}
    ]
  }],
  "response_format": {"type": "json_schema",
                      "json_schema": {"name": "vlmpilot_action", "schema": <S_t>}}
}
\end{lstlisting}

\begin{lstlisting}[style=promptstyle, float=tp, caption={Native Gemini multimodal request construction. Here, \texttt{generation\_config} carries the decoding settings summarized in Section~\ref{app:serving}.}, label={lst:gemini-request}]
generate_content(
    model = <model identifier>,
    contents = [P_user_t, Part(data=JPEG(I_t), mime_type="image/jpeg")],
    config = generation_config   # + system_instruction=P_sys, response_mime_type="application/json", response_schema=S_t
)
\end{lstlisting}

\FloatBarrier
\section{Success Criterion and Metrics}
\label{app:verifier}
\label{app:bench:metrics}

Algorithm~\ref{alg:verdict} states the success criterion.

\begin{algorithm}[h]
\caption{The \bench{} success criterion. The same rule is applied to every drone in every episode, for all task types and for the \emph{single-drone}, \emph{multi-drone} and \emph{commander + sub-agents} settings alike.}
\label{alg:verdict}
\begin{algorithmic}[1]
\Require arrival declarations $D$, each a triple $(t, d, v)$ of time, distance to the
  target and visibility measured against ground truth at $t$, the radius $\delta$ and
  the episode cap $T_{\max}$
\State $D \gets$ entries of $D$ with $t \le T_{\max}$, ordered by $t$
  \Comment{a declaration logged during shutdown does not count}
\For{$(t, d, v) \in D$}
  \If{$d \le \delta$ \textbf{and} $v$}
    \State \Return \textsc{Success}, $\mathrm{NE} \gets d$
    \Comment{the first qualifying declaration decides}
  \EndIf
\EndFor
\State \Return \textsc{Fail}, $\mathrm{NE} \gets d$ at the last action
  \Comment{reported where the episode ended, declared or not}
\end{algorithmic}
\end{algorithm}

Passing through the goal region is not success. An episode with no declaration fails even if the drone came within $\delta$, since crediting proximity would credit models that never declare for flights they did not know had arrived.
The minimum distance and OSR are diagnostics, read from the measurement taken at every step, one per model output.

\paragraph{Measurement.}
Distance is measured to the target centre, and the success radius $\delta$ applies to this distance.
Visibility is the simulator's own detection test, so it accounts for both the view frustum and occlusion.
The median start distance of the \emph{single-drone} episodes is about $16$\,m.
In the \emph{single-drone} runs of Table~\ref{tab:main}, one zero-shot Cosmos3-Edge-2B episode stuck at spawn through a simulator fault. It was re-flown from the same start pose, and the retry replaces it.
Host load can slow the control loop and lower success on moving targets, so runs whose steps had been slowed were re-measured with one simulator per GPU, and the per-step timestamps of the other runs were checked.
With four drones every measurement is threaded by \texttt{vehicle\_name}, so distance, visibility and declarations are recorded per drone.
The verdict pools the four sets of declarations, and the per-drone records supply the per-structure comparison and the coordinate-copying rate of Section~\ref{sec:experiments:org}.
The measurement settings are fixed across the suite.
\begin{center}
\footnotesize
\begin{tabular}{@{}ll@{}}
\toprule
Pose trace & ground-truth poses sampled at $2$\,Hz over its own RPC connection \\
Visibility & engine detections (\texttt{simGetDetections}), filtered to the target mesh name \\
Detection radius & $300$\,m, queried immediately after the image capture \\
Declaration pose & recorded at the \texttt{[FRAME]} the declaration was produced from \\
Success radius & $\delta$ per episode, $5$\,m outdoors and $2.5$\,m indoors \\
\bottomrule
\end{tabular}
\end{center}

\FloatBarrier
\section{Fine-Tuning Cosmos3-Edge-2B}
\label{app:bench:ft}

\paragraph{Data.}
The fine-tuning set contains no frames from the benchmark maps. AirSim is used only for evaluation.
Images come from three public RGB-D datasets with per-pixel ground-truth depth. These are NYU Depth v2~\citep{Silberman:ECCV12} (338 train / 25 val images, indoor, Kinect), SUN RGB-D~\citep{song2015sunrgbd,janoch2011b3do,xiao2013sun3d} (891 / 102, indoor, Kinect) and TartanAir~\citep{tartanair2020iros} (2{,}358 / 192, outdoor, simulated, drone viewpoint), 3{,}587 / 319 images in total (3{,}396 / 318 distinct photographs). SUN RGB-D redistributes the NYU Depth v2 images as its \texttt{kv1/NYUdata} subset (re-cropped and re-encoded), and all 891 SUN RGB-D training images come from this subset, 191 of which also appear in our NYU Depth v2 split.
We match the two releases by NYU frame ID and split at the scene level. No validation image shares an NYU scene with any training image. Each image is augmented into several views (resolution-rescaled and cropped variants, and copies that differ only in the camera parameters written to EXIF), so that one frame yields multiple training samples.
From these views three single-turn tasks are synthesised (Table~\ref{tab:ft:data}). The tasks are point depth (image and pixel coordinate $\to$ depth), referring point and depth (image and object description $\to$ point and depth in the \texttt{go} format), and drone action (image, target description and action history $\to$ one action JSON in the format of Section~\ref{sec:agent:action}).

\begin{table}[!htb]
\centering
\footnotesize
\caption{Fine-tuning rows per task and source.}
\label{tab:ft:data}
\begin{tabular}{@{}lrrrrr@{}}
\toprule
Task & Train & NYU & SUN RGB-D & TartanAir & Val \\
\midrule
Point depth                 & 12{,}530 & 1{,}517 & 3{,}815 &  7{,}198 &   393 \\
Referring point $+$ depth   & 25{,}910 & 3{,}126 & 8{,}161 & 14{,}623 &   732 \\
Drone action                & 13{,}039 & 1{,}462 & 3{,}893 &  7{,}684 & 1{,}372 \\
\midrule
Total                       & 51{,}479 &         &         &          & 2{,}497 \\
\bottomrule
\end{tabular}
\end{table}

\paragraph{Targets and depth labels.}
Up to five objects per image are tagged by GPT-5.5 with a natural-language description and a normalised coordinate. Only opaque surfaces are tagged, and the description must identify the object uniquely.
The depth label is read from the ground-truth depth map at the tagged pixel, snapped to the nearest valid pixel when the sensor value is missing, and taken as the median over a window of $2\%$ of the short image side of the pixels within $\pm5\%$ of the seed depth.
Depth is the planar $z$ along the camera axis.
A second VLM pass checks that the marked point matches the description, and only passing rows are kept.

\begin{figure}[!htb]
\centering
\includegraphics[width=\textwidth]{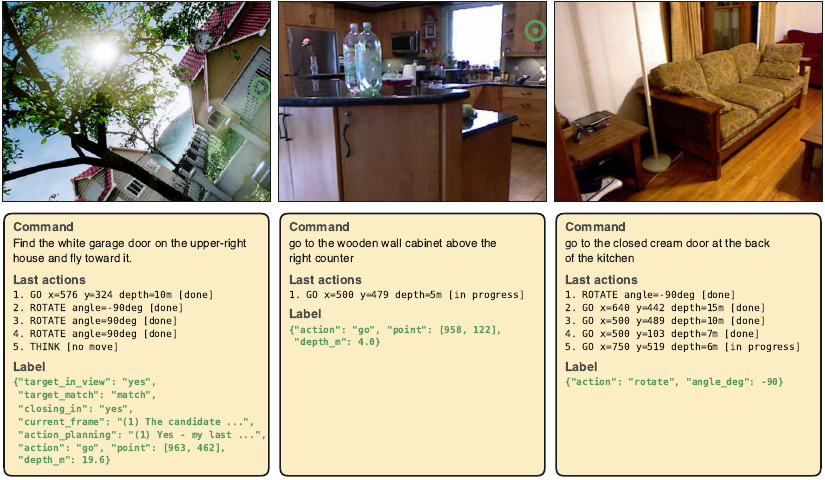}
\caption{Three drone-action rows of the fine-tuning set. Each shows the image, the command and the last actions of the user message, and the label. The first row follows a \texttt{think}, and its two free-text plan fields are cut short here. The ring marks the \texttt{go} point. The system message is the prompt of Listing~\ref{lst:prompt}.}
\label{fig:ft:examples}
\end{figure}

\paragraph{Drone-action labels.}
Actions are assigned by rule (train rows, \texttt{finished} 4{,}322, \texttt{go} 3{,}900, \texttt{rotate} 3{,}614, \texttt{think} 1{,}203).
\texttt{finished} requires ground-truth distance $\le 5$\,m, a target bounding box of at least $5\%$ of the image area and a box bottom in the lower third of the frame, the numeric form of the arrival wording in the prompt. A near target that fails the visual conditions is not labelled \texttt{finished}.
\texttt{go} rows carry the point and depth of a visible target beyond $5$\,m (2{,}430) or of a target cut by the image border in a derived crop (1{,}177).
\texttt{rotate} rows pair an image with a description of an object absent from it, verified absent by a VLM, and label a $\pm90^\circ$ sweep (2{,}925) or a $\pm20$--$25^\circ$ turn (317).
\texttt{think} rows show an image without the target after a history of four or five same-direction \texttt{rotate} actions, that is, after a full turn.
A further 1{,}377 rows prepend the plan-schema fields of Section~\ref{sec:agent:action} (target in view, target match, closing in, current frame, action planning) to a \texttt{finished} (712), \texttt{rotate} (372) or \texttt{go} (293) action, matching the step that follows a \texttt{think} at deployment. No free-text chain of thought is used.
Every drone-action row uses the deployment prompt, with the system message of Listing~\ref{lst:prompt} and a user message that holds the command and the last actions. Figure~\ref{fig:ft:examples} shows a row after a \texttt{think}, whose label carries the plan-schema fields before a \texttt{go}, a plain \texttt{go} row and a \texttt{rotate} row.

\paragraph{Canonical depth.}
The apparent size of an object at a given distance depends on the horizontal field of view $f_x/W$, and a model trained on metric depth learns the camera prior of each dataset. Adding the focal length as text had no effect (identical predictions on $43\%$ of items, zero change in mean relative accuracy), and on unseen cameras the fine-tuned model fell to the accuracy of the base model.
The label is therefore the depth the object would have under a reference camera with $f_x/W = 1$ (horizontal field of view $53^\circ$),
\[
D_c = D \cdot W / f_x , \qquad D = D_c \cdot f_x / W ,
\]
which makes image $\to D_c$ camera-independent, and the metric depth is recovered from the camera parameters after inference.
The agent applies the recovery with $f_x/W = 1/(2\tan(\mathrm{hFOV}/2))$, which is $0.5$ for the $90^\circ$ AirSim camera and $0.93$ for the $56.4^\circ$ Tello camera.

\paragraph{Mixture and optimisation.}
The depth data are mixed at $10\%$ with $90\%$ MAmmoTH-VL instruction data~\citep{guo2024mammothvlelicitingmultimodalreasoning,visualwebinstruct,xu2024llavacot} (Table~\ref{tab:ft:mix}). Over the $400$ training steps this corresponds to about $2.1$ epochs of MAmmoTH and $9.3$ epochs of the depth data.
The base is the released Cosmos3-Edge-2B reasoner VLM~\citep{nvidia2026cosmos3} without further pre-training. The vision encoder, projector and LLM are all fine-tuned.
Training uses $32$ H100 80\,GB GPUs, $768$ micro-batches per step with dynamic batching at 32k tokens per micro-batch and a 32k maximum sequence, AdamW ($\beta = (0.9, 0.95)$, weight decay $0.01$, gradient clipping $1.0$), a learning rate of $5\times10^{-5}$ ($5\times10^{-6}$ for the vision encoder) with $40$ warm-up steps and cosine decay to $5\times10^{-6}$, bf16, and images up to $1920\times1080$.

\begin{table}[!htb]
\centering
\footnotesize
\caption{Sampling mixture. Rows are drawn with the stated probability per dataset.}
\label{tab:ft:mix}
\begin{tabular}{@{}lrr@{}}
\toprule
Dataset & Rows & Share \\
\midrule
MAmmoTH-VL single-image (OneVision stage) & 1{,}023{,}562 & 45.0\% \\
MAmmoTH-VL multi-image (OneVision stage)  &   575{,}080 & 25.3\% \\
MAmmoTH-VL2 (VisualWebInstruct, LLaVA-CoT-100k) & 445{,}601 & 19.6\% \\
Depth (point depth)                        &    12{,}530 &  2.4\% \\
Depth (referring point $+$ depth)          &    25{,}910 &  5.0\% \\
Depth (drone action)                       &    13{,}039 &  2.5\% \\
\bottomrule
\end{tabular}
\end{table}

\paragraph{General capability.}
Table~\ref{tab:ft:general} compares the base and fine-tuned checkpoints on the eight benchmarks the harness runs in direct mode, that is, those whose reasoning budget is too short to carry a chain of thought.
The four benchmarks that the harness runs in reasoning mode, with a budget of 8k to 16k tokens, are outside this comparison.
What the table measures is the effect of the mixture of the previous paragraph, not of the depth data alone. Nine tenths of the fine-tuning tokens are general instruction data, so a rise on these benchmarks is as likely to come from that data as from ours.

\begin{table}[!htb]
\centering
\caption{General capability of Cosmos3-Edge-2B before and after fine-tuning (metric $\times 100$). The fine-tuned column is the $10\%$ depth mixture at $400$ steps described above. Both columns are run in direct mode, since the harness allots these benchmarks $16$ to $256$ reasoning tokens, too few to carry a chain of thought. DocVQA, InfoVQA and TextVQA are validation splits, and the others are the default test split.}
\label{tab:ft:general}
\input{tables/table_general.tex}
\end{table}

\FloatBarrier
\section{Single-Drone Failure Cases}
\label{app:qual}

\begin{figure}[!htb]
\centering
\includegraphics[width=\textwidth]{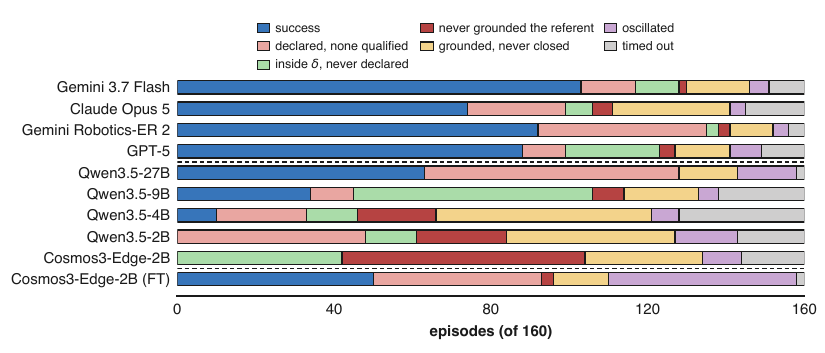}
\caption{Failure modes of the \emph{single-drone} runs, $160$ per model. Success and the six modes partition all runs. Most failures of Gemini Robotics-ER 2 and Qwen3.5-27B are declarations that never qualified, and nearly half of those of Qwen3.5-9B are entries into $\delta$ without a declaration. The other models spread their failures over several modes.}
\label{fig:failmodes}
\end{figure}

Figure~\ref{fig:failmodes} partitions every \emph{single-drone} run into success and six failure modes, labeled from the verdicts alone.
\emph{Declared} means the model declared but no declaration qualified.
\emph{Arrived} means the drone entered $\delta$ and never declared.
\emph{Never grounded} means the target was inside the $45^\circ$ half-FOV in under $10\%$ of steps.
\emph{Oscillated} means the path was over eight times the net displacement, with at least $720^\circ$ of accumulated yaw.
\emph{Grounded, never closed} means the drone faced the target but closed under $30\%$ of the start distance.
\emph{Timed out} covers the rest.

\begin{figure}[!htb]
\centering
\includegraphics[width=0.48\textwidth]{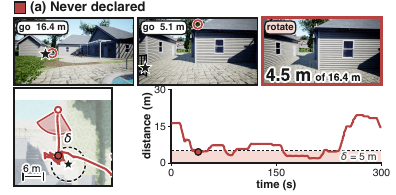}\hfill
\includegraphics[width=0.48\textwidth]{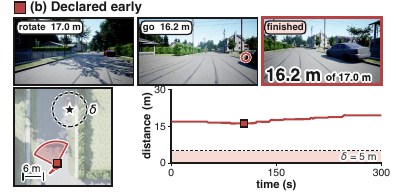}
\caption{The two ways a declaration goes wrong, one episode each on the residential map.
\textbf{(a)} Qwen3.5-9B is inside $\delta$ from step $7$ and never declares in
$83$ steps. \textbf{(b)} Cosmos3-Edge-2B (FT) declares at step $29$, $16.2$\,m out
with the target not in view, and never enters $\delta$. Each strip ends on the
marked frame, and the two windows share one scale.}
\label{fig:examples}
\end{figure}

\begin{figure}[!htb]
\centering
\includegraphics[width=\textwidth]{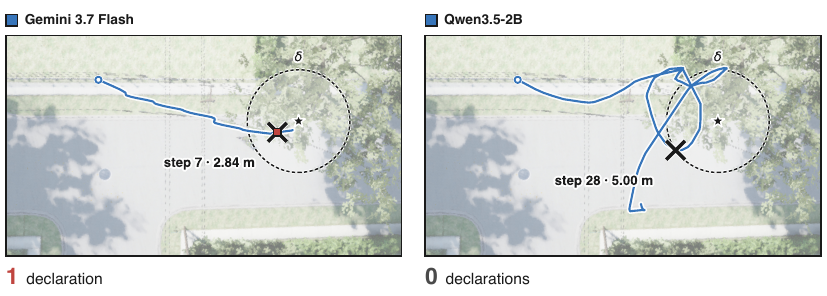}
\caption{The same episode flown twice (residential map, \emph{approach},
\texttt{"go to the cyclist"}), Gemini 3.7 Flash left and Qwen3.5-2B right, on
one plate at one scale. Both start $20$\,m out, and $\times$ marks the first step
inside $\delta$ on each side. The frontier model crosses at step $7$ and declares
at once, $2.8$\,m short. The 2B model crosses at step $28$, comes within
$1.7$\,m, declares nothing, drifts back out of the ring and times out $11$\,m
away.}
\label{fig:qual:pair}
\end{figure}

\begin{figure}[!htb]
\centering
\includegraphics[width=0.48\textwidth]{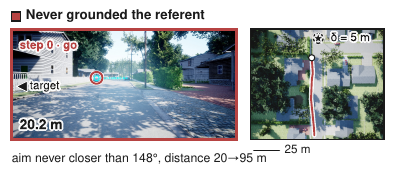}\hfill
\includegraphics[width=0.48\textwidth]{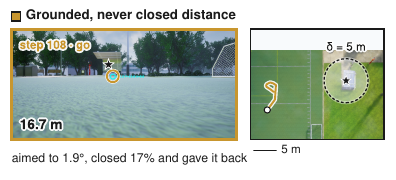}\\[6pt]
\includegraphics[width=0.48\textwidth]{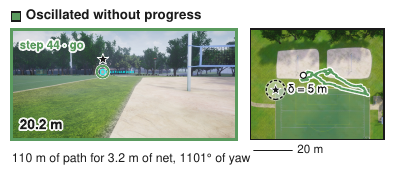}\hfill
\includegraphics[width=0.48\textwidth]{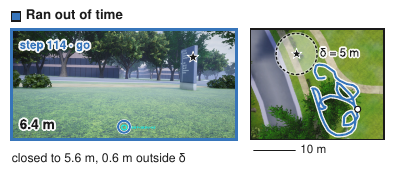}
\caption{Four failure modes, one episode each. Left to right and top
to bottom, Claude Opus 5 never grounds the referent, Qwen3.5-2B faces the target
but never closes, Gemini 3.7 Flash oscillates, and Qwen3.5-2B runs out of time.}
\label{fig:qual:fails}
\end{figure}

The two declaration failures are in Figure~\ref{fig:examples}, and Figure~\ref{fig:qual:pair} flies one episode with a model that declares and one that does not.
Figure~\ref{fig:qual:fails} shows a representative trajectory for each of the four modes that do not involve the declaration.

\FloatBarrier

\FloatBarrier
\section{Additional Results}
\label{app:results}

\begin{figure}[!htb]
\centering
\includegraphics[width=\textwidth]{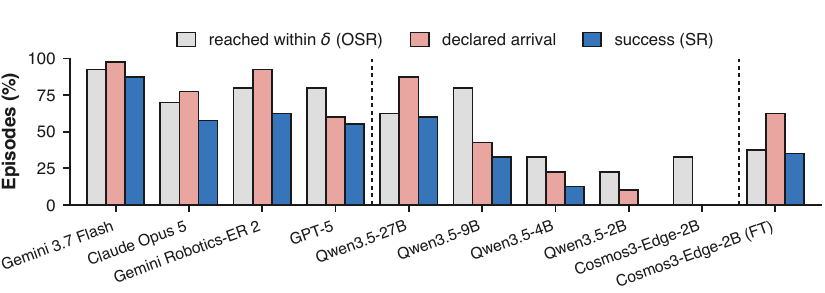}
\caption{Approach decomposed into entering $\delta$, declaring arrival and succeeding, over the $40$ approach runs of each model, so each bar is a multiple of $2.5$ points. Gemini 3.7 Flash, Claude Opus 5 and Gemini Robotics-ER 2 lose most of their episodes by never reaching $\delta$. GPT-5 and Qwen3.5-9B often enter $\delta$ and never declare. Qwen3.5-27B and the fine-tuned model declare more often than they enter $\delta$.}
\label{fig:gap}
\end{figure}

\paragraph{Reaching and declaring.}
Figure~\ref{fig:gap} splits approach into entering $\delta$, declaring arrival and succeeding.

\paragraph{Policy switches.}
Table~\ref{tab:ablation} reports two switches.
\texttt{think} off removes \texttt{think} from the prompt and the action set, so the model cannot call it.
Visual history keeps the text history of the last five actions and adds the view the drone had when it chose each one, at half the width of the current frame.

\begin{figure}[!htb]
\centering
\includegraphics[width=\textwidth]{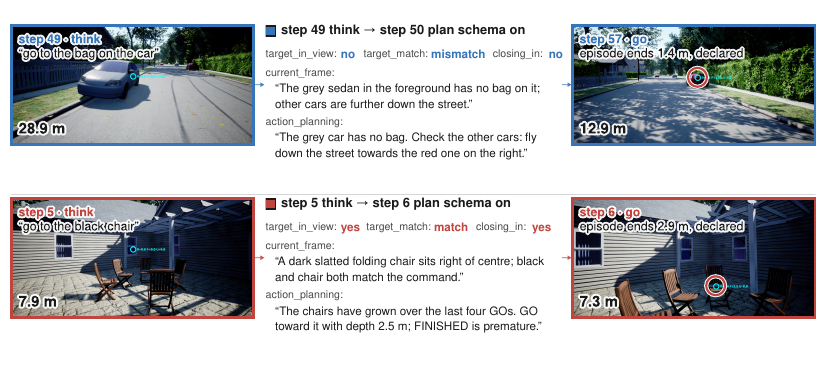}
\caption{One \texttt{think} step in two episodes. Left, the frame that
triggered it. Centre, the plan schema the next call returned, field by field.
Right, a later frame with the model's emitted point.
Above, Gemini Robotics-ER 2 closing on the wrong car. The schema records no
target in view and a mismatch, redirects to the red car down the street, and the
episode ends declared at $1.4$\,m. Below, Claude Opus 5 with the chair already
in view. The schema confirms what the frame shows and the plan restates the
\texttt{go} under way.}
\label{fig:qual:think}
\end{figure}

\paragraph{\texttt{think}.}
\texttt{think} lets the model deliberate only at the steps it chooses, instead of spending decoding time at every step.
In some episodes this helps. In the upper episode of Figure~\ref{fig:qual:think}, the model notices that the drone is approaching the wrong car and turns it toward the target.
However, the drone holds position until the plan step returns, so the next movement is delayed by a factor of two or more, and a moving target drifts away during this time.
Without \texttt{think}, SR on track and search + track rises for eight of the ten models, from $29.1\%$ to $37.8\%$ in total, and mean SR rises for six models.
The switch also replaces the prompt's instruction to think when stuck with an instruction to rotate or move, so part of this gain may come from the prompt change.

\begin{table}[!htb]
\centering
\caption{Two policy switches on the \emph{single-drone} suite, mean SR over the four task types (\%). Baseline is the Table~\ref{tab:main} setting, 20 episodes per cell flown twice. \texttt{think} off removes \texttt{think} from the prompt and the output schema. Visual history adds the views of the last five steps, at half width, to the text history. Each switch reruns the suite once per model (20 per cell). The last row averages the ten models.}
\label{tab:ablation}
\input{tables/table_ablation.tex}
\end{table}

\paragraph{Visual history.}
Only the fine-tuned model clearly benefits from visual history. Its SR rises from $31.2\%$ to $43.8\%$, mostly on static targets.
The zero-shot models lose on moving targets. Their SR on track and search + track falls from $29.3\%$ to $21.7\%$, and Qwen3.5-27B drops to zero (Table~\ref{tab:mmhist}).
On static targets the zero-shot models lose about as much as in the \texttt{think}-off run, which adds no past views, so this loss cannot be attributed to the views.
The added views also increase the per-step latency of every model except Claude Opus 5, by up to a factor of $1.8$.

\begin{table}[!htb]
\centering
\caption{The visual-history arm decomposed (\%, except latency in seconds and declarations per episode). Base is the Table~\ref{tab:main} setting over 160 runs per model. Hist.\ is the visual-history arm of Table~\ref{tab:ablation}. Latency is the median model time per step. Static covers approach and search, moving covers track and search + track.}
\label{tab:mmhist}
\input{tables/table_mmhist.tex}
\end{table}

\paragraph{Run-to-run variance.}
Table~\ref{tab:main} pools two repetitions of the suite under identical settings, and Table~\ref{tab:variance} lists them separately.
With 40 binary runs per cell, a perfectly stable pipeline still has a standard deviation of 6 to 8 points per cell.
The two repetitions differ by 5.0 points on average across the 40 model and task cells, and by at most 20, which is within this range.
The tiers of Table~\ref{tab:main} are therefore stable, but the order of neighbouring models within a tier is within the noise.

\begin{table}[!htb]
\centering
\caption{The two repetitions of the \emph{single-drone} suite behind Table~\ref{tab:main}, SR (\%) over 20 episodes per task and 80 in total (first/second repetition, both maps pooled).}
\label{tab:variance}
\setlength{\tabcolsep}{5pt}
\renewcommand{\arraystretch}{1.12}
\footnotesize
\begin{tabular}{@{}lccccc@{}}
\toprule
& \multicolumn{5}{c}{SR (\%)} \\
\cmidrule(lr){2-6}
Model & Approach & Search & Track & Search + Track & Total \\
\midrule
Gemini 3.7 Flash     & 90/85 & 60/75 & 60/50 & 40/55 & 62.5/66.2 \\
Gemini Robotics-ER 2 & 65/60 & 55/40 & 70/70 & 45/55 & 58.8/56.2 \\
GPT-5                & 50/60 & 35/40 & 60/65 & 65/65 & 52.5/57.5 \\
Claude Opus 5        & 55/60 & 45/50 & 35/45 & 50/30 & 46.2/46.2 \\
\midrule
Qwen3.5-27B          & 60/60 & 45/45 & 25/25 & 20/35 & 37.5/41.2 \\
Cosmos3-Edge-2B (FT) & 35/35 & 30/40 & 25/30 & 25/30 & 28.8/33.8 \\
Qwen3.5-9B           & 35/30 & 15/15 & 35/30 & 5/5 & 22.5/20 \\
Qwen3.5-4B           & 10/15 & 10/0 & 10/5 & 0/0 & 7.5/5 \\
Qwen3.5-2B           & 0/0 & 0/0 & 0/0 & 0/0 & 0/0 \\
Cosmos3-Edge-2B      & 0/0 & 0/0 & 0/0 & 0/0 & 0/0 \\
\bottomrule
\end{tabular}
\end{table}

\FloatBarrier
\section{Static Benchmarks}
\label{app:static}

Table~\ref{tab:static} places the instruction-following and general VLM benchmarks next to the flight results of every model, and Table~\ref{tab:static:corr} gives their rank correlations with SR, OSR and conversion.

SysBench is the benchmark most closely related to the declaration.
Its rank correlation with conversion is $0.95$, while that with OSR is $0.62$.
Ordering the nine off-the-shelf models by SysBench gives nearly the same order as conversion, except that Qwen3.5-27B moves from third to first.
For example, Qwen3.5-9B reaches $\delta$ more often than Qwen3.5-27B, but converts about one third of those episodes into a success, whereas Qwen3.5-27B converts almost all of them.
The two models differ by less than four points on IFEval and MM-IFEval but by nine points on SysBench.

This may be because the declaration rule is given in the system message.
The condition for \texttt{finished} is stated there (Listing~\ref{lst:prompt}), and the model has to apply it to a new frame at every step.
SysBench measures adherence to constraints given in the system message.
IFEval and MM-IFEval measure constraints on a single response, and like the general benchmark MME they correlate similarly with SR, OSR and conversion.
VSysBench also measures system-message compliance, but with image inputs, and it correlates more with OSR than with conversion. Its score probably also reflects visual recognition.

The fine-tuned model is an exception.
It has the lowest SysBench score, but converts almost every episode that enters $\delta$ into a success.
Its drone-action data label \texttt{finished} by a fixed rule (Appendix~\ref{app:bench:ft}), so the declaration was learned directly rather than through system-prompt following.
The correlations are computed over only nine models, and SysBench is in Chinese while our prompt is in English.

\begin{table}[!htb]
\centering
\caption{Static benchmarks against flight, over the roster of Table~\ref{tab:models}. SR and OSR are the means of Table~\ref{tab:main}. Conversion is SR divided by OSR, the share of the episodes that entered $\delta$ that also succeeded. The frontier models were run in reasoning mode and the open models in direct mode. The benchmarks are SysBench~\citep{qin2024sysbench}, VSysBench~\citep{yeo2026compliance}, IFEval~\citep{zhou2023instruction}, MM-IFEval~\citep{ding2025mmifengine}, MME~\citep{fu2026mme} and MMStar~\citep{chen2024we}. SysBench is in Chinese and judged by GPT-4o, VSysBench in English and judged by GPT-5-mini, while the prompts of this paper are in English.}
\label{tab:static}
\input{tables/table_static.tex}
\end{table}

\begin{table}[!htb]
\centering
\caption{Spearman rank correlations between the columns of Table~\ref{tab:static}, over the nine off-the-shelf models. SysBench correlates with conversion far more than with OSR. For the other benchmarks the gap is small or reversed.}
\label{tab:static:corr}
\input{tables/table_static_corr.tex}
\end{table}

\FloatBarrier
\section{Isaac Sim}
\label{app:isaac}

We also run a smaller \emph{single-drone} suite in Isaac Sim~\citep{gao2026nvidia} with the PX4 flight stack, on one large outdoor street scene, Brownstone.
It has $28$ episodes, eight approach, eight search, six track and six search + track, each flown once.
The prompt, the actions, the criterion with $\delta = 5$\,m and the $300$\,s cap of simulated time are those of the AirSim suite.
A scripted oracle reaches the target in all $28$ episodes.

Table~\ref{tab:isaac} reports the results.
The ranking broadly follows AirSim (Spearman $\rho = 0.85$ over the ten models).
The four frontier models lead in both simulators, and Qwen3.5-2B and zero-shot Cosmos3-Edge-2B succeed in no episode in either.
The largest departure is Qwen3.5-27B, which declares in $27$ of the $28$ episodes but outside $\delta$ in $23$ of them.

The order also shifts elsewhere, and two differences between the simulators likely contribute.
First, the scene is large and runs with collision checking, so one second of simulated time takes about eleven seconds of wall-clock time.
A four-second model call then lets the target move for only about $0.4$\,s, and the latency cost on moving targets (Section~\ref{sec:experiments:sr}) is much smaller than in AirSim.
This also removes the edge of the fast models. The fine-tuned 2B model, which leads on hardware through its short step latency, falls to $18\%$, while Claude Opus 5 rises from fourth to second.
Second, the rendered images differ from AirSim's in quality.

\begin{table}[!htb]
\centering
\caption{Single-drone results in Isaac Sim (\%, $28$ episodes on one outdoor scene, flown once). The last column is the AirSim mean SR of Table~\ref{tab:main}.}
\label{tab:isaac}
\input{tables/table_isaac.tex}
\end{table}

\FloatBarrier
\section{Four-Drone Details}
\label{app:multidrone}

\subsection{Setup}
The two command structures are those of Section~\ref{sec:agent:org}, and the three categories are those of Appendix~\ref{app:bench} (Figure~\ref{fig:multi:layout}), with ten episodes each.
The sub-agents are Gemini Robotics-ER~2, and each episode is cut at 300~s.
An order lives for six sub-agent steps, and a drone whose order has expired holds position until the commander issues a new one, so a commander call that outlasts the orders stalls the fleet.
We call the text-free top-down map that the commander receives the control-tower map.
In Three-sign sum, the goal sign is the small sign at the end of the corridor whose entrance shows the sum.

\subsection{Why Commanding Helps}
The advantage of commander + sub-agents over multi-drone comes from dividing the work between the commander and the sub-agents.
With Gemini 3.7 Flash as the model, all structures solve License plate matching, whereas on Three-sign sum only a commander that keeps issuing commands achieves a high success rate (Table~\ref{tab:multi:struct}).
In the \emph{one-shot} control, the commander issues a single command per drone at take-off. In the \emph{no commander} control, each sub-agent acts on the instruction alone.

\begin{table}[!htb]
\centering
\caption{Command structure with Gemini 3.7 Flash as the model. Success rate over ten episodes per category. The last two columns give how many times per minute the fleet moves and Closest, the smallest distance between any drone and the goal sign in an episode, both as medians over the ten Three-sign sum episodes. They were not recorded for one-shot.}
\label{tab:multi:struct}
\input{tables/table_multi_struct.tex}
\end{table}

\subsection{What a Good Commander Does}
\begin{figure}[!htb]
\centering
\includegraphics[width=\textwidth]{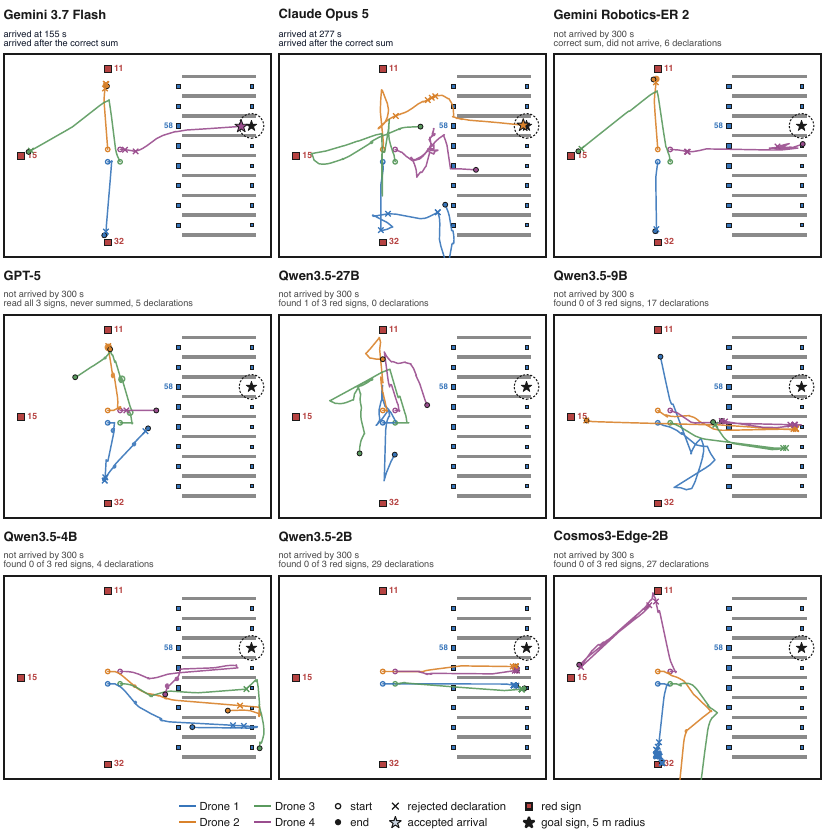}
\caption{The same Three-sign sum episode flown with each of the nine zero-shot commanders under \emph{commander + sub-agents} with the map. The line under each name is the judge's reading of the episode. Each colour is one drone. Crosses are arrival declarations the judge rejected, and the filled star is the accepted one.}
\label{fig:multi:models}
\end{figure}

On Three-sign sum, successful commanders gather all three numbers and compute their sum before sending a drone into a corridor.
When the nine zero-shot commanders fly the same episode with the same sub-agents, only two of them follow this strategy (Figure~\ref{fig:multi:models}).
Gemini Robotics-ER~2 obtains the correct sum but fails to bring a drone to the goal sign. GPT-5 reads all three signs without adding them, and the smaller models fail to locate the red signs.
The fine-tuned Cosmos commander is excluded because it typically returns a single command for the whole fleet.

Unlike the single-drone agent, whose reasoning is kept at the lowest level (Appendix~\ref{app:serving}), the Gemini 3.7 Flash, Claude Opus 5 and GPT-5 commanders of Section~\ref{sec:experiments:org} run with reasoning on, at the highest level whose call returns within the lifetime of an order (Appendix~\ref{app:serving}).
With reasoning disabled, the success rate of Claude Opus 5 on Three-sign sum falls from 80\% to 40\%, while the two easier categories do not change.
Even so, its commander total of 73.3\% stays above its multi-drone total of 46.7\%, so the gain of Section~\ref{sec:experiments:org} does not come from reasoning alone.
Gemini 3.7 Flash cannot disable reasoning, and even at a zero budget it keeps about 220 reasoning tokens per call and its success rate does not fall.

The control-tower map helps in the same place (Table~\ref{tab:multi:map}).
Across three commanders, the success rate on Three-sign sum increases from 23\% to 57\% with the map, while the two easier categories change little.

As a sub-agent, the fine-tuned Cosmos model shows the opposite of its result as a commander.
With Gemini 3.7 Flash commanding, fine-tuned Cosmos sub-agents raise success over zero-shot Cosmos sub-agents from 40\% to 90\% on License plate matching and from 0\% to 50\% on Sign matching, though neither succeeds on Three-sign sum, so the drone data teach the model to follow commands but not to give them.

\begin{table}[!htb]
\centering
\caption{Success rate with and without the control-tower map, for three commanders over ten episodes per category and for all three over 30.}
\label{tab:multi:map}
\input{tables/table_multi_map.tex}

\end{table}

\subsection{Real Flights}
The real flights use two Tello drones on Three-sign sum without the map (Appendix~\ref{app:bench:real}).
A set denotes one arrangement of the three numbers and the four numbered objects in the room.
In set 5 (Figure~\ref{fig:real}(b)), one drone reads 5 and the other reads 20 and 35.
The commander computes the sum and sends the first drone to chair 60.

\FloatBarrier
\input{content/related.tex}

%% file: tables/table_models.tex
\setlength{\tabcolsep}{3.5pt}
\renewcommand{\arraystretch}{1.12}
\footnotesize
\begin{tabular}{@{}lll@{}}
\toprule
Model & Identifier & Access \\
\midrule
Gemini 3.7 Flash~\citep{deepmind2026gemini37flash} & \texttt{gemini-3.7-flash} & Google API \\
Claude Opus 5~\citep{anthropic2026opus5} & \texttt{anthropic/claude-opus-5} & OpenRouter \\
Gemini Robotics-ER 2~\citep{deepmind2026roboticser2} & \texttt{gemini-robotics-er-2-preview} & Google API \\
GPT-5~\citep{openai2025gpt5} & \texttt{gpt-5} & OpenAI API \\
\midrule
Qwen3.5-27B~\citep{qwen35vl} & \texttt{qwen/qwen3.5-27b} & OpenRouter \\
Qwen3.5-9B & \texttt{qwen/qwen3.5-9b} & OpenRouter \\
Qwen3.5-4B & \texttt{Qwen/Qwen3.5-4B} & local vLLM \\
Qwen3.5-2B & \texttt{Qwen/Qwen3.5-2B} & local vLLM \\
Cosmos3-Edge-2B~\citep{nvidia2026cosmos3} & \texttt{nvidia/Cosmos3-Edge} & local vLLM \\
\midrule
Cosmos3-Edge-2B (FT, ours) & in-house checkpoint, Appendix~\ref{app:bench:ft} & local vLLM \\
\bottomrule
\end{tabular}

%% file: content/prompt.tex
\begin{lstlisting}[style=promptstyle, float=tp, caption={The system prompt, verbatim. Long lines are wrapped for the column, and the content is unaltered. Braces around \texttt{action} are literal JSON in the emitted prompt.}, label={lst:prompt}]
You are an autonomous drone pilot in a simulator. You see ONE egocentric
image from the drone's forward camera. Your job: reach the target described below,
then declare arrival. Decide WHERE the target is, then output ONE action.

Command: {instruction}

## Output - EXACTLY ONE action as a single JSON object (choose one of four):

1) GO - fly toward a point you can see. Use when the target, or the path toward it, is visible.
   {"action":"go","point":{point_fmt},"depth_m":<meters>}
   - point  : on the target (or along the way to it). x: 0-1000 left->right (500=center).
              y: 0-1000 top->bottom (500=center).
              The image you receive is 640x360 px. Do NOT answer in pixels -
              scale x by 1000/640 and y by 1000/360. A spot a quarter in from
              the left and a quarter down from the top is pixel (160, 90)
              and is 250 on BOTH axes. The axes scale differently, so 500,500 is the exact
              centre of the frame - do not fall back on it unless the target is centred.
   - depth_m: how far AHEAD to fly, in METERS - the distance along the direction the
              camera is facing, NOT the diagonal distance to the point.
              FAR target -> LARGE depth (take big strides). CLOSE target -> SMALL depth (creep in carefully).

2) ROTATE - yaw in place (hold position and altitude), turning the camera to look around.
   Use to SEARCH when the target is NOT visible.
   {"action":"rotate","angle_deg":<deg>}
   - angle_deg: + = turn right, - = turn left, up to +-90 per step.
     Keep sweeping the SAME direction across steps so you cover a full 360, don't oscillate.

3) THINK - hold position and spend this step thinking. Use when the view is unclear, there are
   multiple candidates, or you are unsure what to do next. The step right AFTER a THINK you will
   reason at length before acting, so THINK BUYS you a careful decision - it is not a wasted step.
   {"action":"think"}

4) FINISHED - declare you have ARRIVED and hold. Use ONLY when you are within a few meters.
   How to tell you are close enough: the target must FILL a large part of the frame. If it still
   fits inside the view with empty margin all around it, you are more than 10 m away - keep going.
   For an object on the ground, its base should reach the lower third of the image.
   {"action":"finished"}

## How to decide

- Can't see the target -> ROTATE to search. Keep sweeping the SAME direction, don't oscillate.
- Rotated a full turn and still nothing -> STOP rotating. More rotation from the same spot shows
  you the same view. GO to a new vantage point (an open direction), then look again from there.
- Target visible but not there yet -> GO toward it. It must look BIGGER and more centered than
  the last step; SHRINK depth_m as it grows. If it isn't getting bigger, you are not closing in.
- Target FILLS much of the frame, centered, clear path -> FINISHED. If there is still empty
  margin around it, you are too far - GO, don't declare.
- Stuck or unsure - view unclear, multiple candidates, or your recent GOs aren't getting you
  closer (circling) -> THINK. The step right after a THINK, reason carefully, then commit.

## Your last {history_len} actions (use them to avoid circling / repeating a mistake)
{history}

Output ONLY the single JSON object. No text outside it.
\end{lstlisting}

%% file: tables/table_general.tex
{\footnotesize
\setlength{\tabcolsep}{6pt}
\renewcommand{\arraystretch}{1.12}
\begin{tabular}{@{}llrrr@{}}
\toprule
Benchmark & Metric & Base & FT & $\Delta$ \\
\midrule
DocVQA (val)        & ANLS         & 75.8 & 85.4 & $+9.6$ \\
ChartQA             & exact match  & 76.9 & 82.5 & $+5.6$ \\
InfoVQA (val)       & ANLS         & 51.9 & 55.1 & $+3.2$ \\
AI2D                & exact match  & 75.5 & 77.1 & $+1.6$ \\
OCRBench            & string match & 75.2 & 74.1 & $-1.1$ \\
TextVQA (val)       & VQA eval     & 80.6 & 79.4 & $-1.2$ \\
HallusionBench      & judge      & 39.2 & 37.1 & $-2.1$ \\
ScienceQA           & exact match  & 96.9 & 90.4 & $-6.5$ \\
\midrule
Mean                &              & 71.5 & 72.6 & $+1.1$ \\
\bottomrule
\end{tabular}}

%% file: tables/table_ablation.tex
{\footnotesize
\setlength{\tabcolsep}{6pt}
\renewcommand{\arraystretch}{1.15}
\begin{tabular}{@{}l>{\centering\arraybackslash}p{4.2em}>{\centering\arraybackslash}p{4.2em}>{\centering\arraybackslash}p{4.8em}@{}}
\toprule
Model & Baseline & \texttt{think} off & Visual history \\
& SR & SR & SR \\
\midrule
Gemini 3.7 Flash & 64.4 & 63.8 & 48.8 \\
Claude Opus 5 & 46.2 & 47.5 & 38.8 \\
Gemini Robotics-ER 2 & 57.5 & 65.0 & 58.8 \\
GPT-5 & 55.0 & 50.0 & 46.2 \\
\midrule
Qwen3.5-27B & 39.4 & 40.0 & 20.0 \\
Qwen3.5-9B & 21.2 & 31.2 & 10.0 \\
Qwen3.5-4B & 6.2 & 8.8 & 8.8 \\
Qwen3.5-2B & 0 & 6.2 & 2.5 \\
Cosmos3-Edge-2B & 0 & 0 & 1.2 \\
\midrule
Cosmos3-Edge-2B (FT) & 31.2 & 28.8 & 43.8 \\
\midrule
Mean & 32.1 & 34.1 & 27.9 \\
\bottomrule
\end{tabular}}

%% file: tables/table_mmhist.tex
{\footnotesize
\setlength{\tabcolsep}{3.6pt}
\renewcommand{\arraystretch}{1.12}
\begin{tabular}{@{}lrrrrrrrrrr@{}}
\toprule
& \multicolumn{2}{c}{Latency (s)} & \multicolumn{2}{c}{Decl./ep} & \multicolumn{2}{c}{OSR} & \multicolumn{2}{c}{SR static} & \multicolumn{2}{c}{SR moving} \\
\cmidrule(lr){2-3}\cmidrule(lr){4-5}\cmidrule(lr){6-7}\cmidrule(lr){8-9}\cmidrule(lr){10-11}
Model & Base & Hist. & Base & Hist. & Base & Hist. & Base & Hist. & Base & Hist. \\
\midrule
Gemini 3.7 Flash      & 3.5 & 4.1 &  2.5 &  4.1 & 72.5 & 52.5 & 77.5 & 57.5 & 51.2 & 40.0 \\
Claude Opus 5         & 2.9 & 2.9 &  5.7 &  9.2 & 51.2 & 49.4 & 52.5 & 42.5 & 40.0 & 35.9 \\
Gemini Robotics-ER 2  & 2.0 & 2.9 &  6.7 &  3.3 & 65.0 & 67.1 & 55.0 & 67.5 & 60.0 & 51.3 \\
GPT-5                 & 1.4 & 2.0 &  1.5 &  2.7 & 71.2 & 67.5 & 46.2 & 35.0 & 63.8 & 57.5 \\
\midrule
Qwen3.5-27B           & 1.0 & 1.6 & 39.4 & 26.8 & 40.6 & 22.5 & 52.5 & 40.0 & 26.2 &  0.0 \\
Qwen3.5-9B            & 1.4 & 2.5 &  0.5 &  1.0 & 63.1 & 32.9 & 23.8 & 15.0 & 18.8 &  5.1 \\
Qwen3.5-4B            & 1.0 & 1.2 &  0.8 &  5.7 & 18.8 & 16.2 &  8.8 & 17.5 &  3.8 &  0.0 \\
Qwen3.5-2B            & 0.8 & 1.3 &  0.5 &  4.6 & 13.8 & 25.0 &  0.0 &  0.0 &  0.0 &  5.0 \\
Cosmos3-Edge-2B       & 0.6 & 0.7 &  0.0 & 16.3 & 26.2 & 21.2 &  0.0 &  0.0 &  0.0 &  2.5 \\
\midrule
Cosmos3-Edge-2B (FT)  & 0.6 & 1.1 & 12.3 &  6.9 & 32.5 & 52.5 & 35.0 & 55.0 & 27.5 & 32.5 \\
\bottomrule
\end{tabular}}

%% file: tables/table_static.tex
{\footnotesize
\setlength{\tabcolsep}{4pt}
\renewcommand{\arraystretch}{1.12}
\begin{tabular}{@{}lrrr rrrr rr@{}}
\toprule
& \multicolumn{3}{c}{Flight} & \multicolumn{4}{c}{Instruction following} & \multicolumn{2}{c}{General} \\
\cmidrule(lr){2-4}\cmidrule(lr){5-8}\cmidrule(lr){9-10}
Model & SR & OSR & Conv. & SysB. & VSysB. & IFEval & MM-IFE. & MME & MMStar \\
\midrule
Gemini 3.7 Flash     & 64.4 & 72.5 & 88.8 & 94.3 & 57.1 & 95.4 & 81.7 & 2626.5 & 78.7 \\
Claude Opus 5        & 46.2 & 51.2 & 90.2 & 95.6 & 54.6 & 90.6 & 79.4 & 2541.9 & 76.3 \\
Gemini Robotics-ER 2 & 57.5 & 65.0 & 88.5 & 92.1 & 49.3 & 93.2 & 76.4 & 2586.2 & 58.2 \\
GPT-5                & 55.0 & 71.2 & 77.2 & 91.1 & 59.5 & 85.2 & 74.3 & 2239.7 & 79.0 \\
\midrule
Qwen3.5-27B          & 39.4 & 40.6 & 96.9 & 92.5 & 30.8 & 87.8 & 67.4 & 2439.2 & 75.9 \\
Qwen3.5-9B           & 21.2 & 63.1 & 33.7 & 83.5 & 25.6 & 84.3 & 64.7 & 2195.5 & 66.4 \\
Qwen3.5-4B           &  6.2 & 18.8 & 33.3 & 77.6 & 22.2 & 81.1 & 62.8 & 2064.5 & 57.7 \\
Qwen3.5-2B           &  0.0 & 13.8 &  0.0 & 50.6 & 20.1 & 66.9 & 47.3 & 1964.3 & 72.4 \\
Cosmos3-Edge-2B      &  0.0 & 26.2 &  0.0 & 15.0 &  9.4 & 68.4 & 41.2 & 1688.7 & 70.9 \\
\midrule
Cosmos3-Edge-2B (FT) & 31.2 & 32.5 & 96.2 &  6.5 &  8.2 & 55.6 & 33.7 & 1670.1 & 66.1 \\
\bottomrule
\end{tabular}}

%% file: tables/table_static_corr.tex
{\footnotesize
\setlength{\tabcolsep}{8pt}
\renewcommand{\arraystretch}{1.12}
\begin{tabular}{@{}lrrr@{}}
\toprule
Benchmark & SR & OSR & Conversion \\
\midrule
SysBench   & $+0.81$ & $+0.62$ & $\mathbf{+0.95}$ \\
VSysBench  & $+0.91$ & $+0.85$ & $+0.71$ \\
IFEval     & $+0.95$ & $+0.80$ & $+0.85$ \\
MM-IFEval  & $+0.95$ & $+0.80$ & $+0.81$ \\
\midrule
MME        & $+0.95$ & $+0.77$ & $+0.85$ \\
MMStar     & $+0.44$ & $+0.48$ & $+0.44$ \\
\bottomrule
\end{tabular}}

%% file: tables/table_isaac.tex
{\footnotesize
\setlength{\tabcolsep}{4pt}
\renewcommand{\arraystretch}{1.05}
\begin{tabular}{@{}l>{\centering\arraybackslash}p{5.2em}>{\centering\arraybackslash}p{4.4em}>{\centering\arraybackslash}p{4.2em}>{\centering\arraybackslash}p{8.6em}>{\centering\arraybackslash}p{3.0em}@{\hspace{10pt}}>{\centering\arraybackslash}p{3.2em}@{}}
\toprule
& \multicolumn{5}{c}{Isaac Sim} & AirSim \\
\cmidrule(lr){2-6}\cmidrule(lr){7-7}
Model & Approach & Search & Track & Search + Track & Mean & Mean \\
\midrule
Gemini 3.7 Flash     & \textbf{100} & 50 & \textbf{67} & \textbf{83} & \textbf{75} & \textbf{64.4} \\
Claude Opus 5        & 88 & 63 & \textbf{67} & 67 & 71 & 46.2 \\
Gemini Robotics-ER 2 & 88 & \textbf{75} & 50 & 33 & 64 & 57.5 \\
GPT-5                & 63 & 38 & 50 & \textbf{83} & 57 & 55.0 \\
\midrule
Qwen3.5-27B          & 13 & 13 & 0 & 33 & 14 & 39.4 \\
Qwen3.5-9B           & 50 & 38 & 50 & 50 & 46 & 21.2 \\
Qwen3.5-4B           & 25 & 50 & 50 & 17 & 36 & 6.2 \\
Qwen3.5-2B           & 0 & 0 & 0 & 0 & 0 & 0 \\
Cosmos3-Edge-2B      & 0 & 0 & 0 & 0 & 0 & 0 \\
\midrule
Cosmos3-Edge-2B (FT) & 25 & 13 & 17 & 17 & 18 & 31.2 \\
\bottomrule
\end{tabular}}

%% file: tables/table_multi_struct.tex
\setlength{\tabcolsep}{4pt}
\renewcommand{\arraystretch}{1.12}
\footnotesize
\begin{tabular}{@{}lccc@{\hspace{12pt}}cc@{}}
\toprule
& \multicolumn{3}{c}{SR (\%)} & \multicolumn{2}{c}{Three-sign sum} \\
\cmidrule(lr){2-4}\cmidrule(l){5-6}
Structure & \shortstack[c]{License plate\\matching} & \shortstack[c]{Sign\\matching} & \shortstack[c]{Three-sign\\sum} & \shortstack[c]{Actions\\per minute} & \shortstack[c]{Closest\\(m)} \\
\midrule
Commander + sub-agents & 100 & \textbf{100} & \textbf{80} & 29 & 3.5 \\
One-shot               & 100 & 80  & 0  &    &     \\
No commander           & 100 & 60  & 10 & 74 & 8.7 \\
Multi-drone            & 100 & 50  & 0  & 8  & 53  \\
\bottomrule
\end{tabular}

%% file: tables/table_multi_map.tex
\setlength{\tabcolsep}{5pt}
\renewcommand{\arraystretch}{1.12}
\footnotesize
\begin{tabular}{@{}lcccccc@{}}
\toprule
SR (\%) & \multicolumn{2}{c}{\shortstack[c]{License plate\\matching}} & \multicolumn{2}{c}{\shortstack[c]{Sign\\matching}} & \multicolumn{2}{c}{\shortstack[c]{Three-sign\\sum}} \\
\cmidrule(lr){2-3}\cmidrule(lr){4-5}\cmidrule(lr){6-7}
Commander & No map & Map & No map & Map & No map & Map \\
\midrule
Gemini 3.7 Flash & 100 & 100 & 90  & 100 & 30 & 80 \\
Claude Opus 5    & 100 & 90  & 100 & 90  & 40 & 80 \\
Qwen3.5-9B       & 80  & 90  & 20  & 30  & 0  & 10 \\
\midrule
All three        & 93  & 93  & 70  & 73  & 23 & 57 \\
\bottomrule
\end{tabular}

%% file: content/related.tex
\section{Related Work}
\label{app:related}
\label{sec:related}
\label{app:coverage}

\paragraph{MLLM-based drone control.}
\label{sec:related:systems}
Several systems control drones with off-the-shelf MLLMs and no task-specific training.
TypeFly~\citep{chen2023typefly} plans over detections. PIVOT~\citep{nasiriany2024pivot} selects among candidate points drawn on the image. SPF~\citep{hu2025spf} emits an image point and a travel distance. Fly0~\citep{xu2026fly0} emits a grounding tuple at about $0.5$\,Hz and delegates trajectories to a $50$\,Hz planner. OnFly~\citep{zheng2026onfly} verifies the model's proposals before execution. A PX4 stack pairs a text LLM with a vision-language model~\citep{lim2025px4dialogue,chen2026airhunt}.
In each of these the model fills one role and a planner, verifier or monitor makes the remaining decisions, so their evaluations compare pipelines rather than models. OnFly uses one 4B model for all baselines, Fly0 sweeps backbones for the grounding role only and finds four within $1.1$ points, and the PX4 stack ranks pairings within one pipeline.
SPF is the closest in task coverage. It evaluates search and follow tasks with six VLMs and flies eleven tasks on a real drone, but the episode is terminated by an external rule on the final position and view, so whether the model knows it has arrived is not measured, and its tasks are saturated at $100\%$ success for the strongest models.
We instead evaluate the model as the sole decision maker, with no planner or verifier in the loop, and require it to declare arrival itself.

\paragraph{Benchmarks for embodied UAV agents.}
\label{sec:related:benchmarks}
Table~\ref{tab:coverage} places the benchmarks closest to ours on seven axes.
Closed-loop control is evaluated by most of them, but each covers part of the task space. DeTrack~\citep{hu2026detrack} and UAV-Track VLA~\citep{zhang2026uavtrackvla} cover moving targets, UAV-ON~\citep{xiao2025uavon}, ESARBench~\citep{zhang2026esarbench} and ActiveFly-Bench~\citep{zhang2026activefly} out-of-view search, and only SPF covers both.
AerialVLN~\citep{liu2023aerialvln} and CognitiveDrone~\citep{lykov2025cognitivedrone} evaluate instruction following along a given route.
MissionBench~\citep{navaratnarajah2026zero} is concurrent work that evaluates MLLMs as the sole decision maker on mission-level tasks in simulation, with targets at least partially visible from the start, static, and approached by one drone. Its agent design, transplanted onto our benchmark, solves 13 of 40 single-drone episodes against 22 for ours, and the gap lies in the search and tracking tasks it does not cover (Table~\ref{tab:mbagent}).
BEDI~\citep{guo2026bedi} scores sub-skills mainly by multiple choice, and UrbanVideo-Bench~\citep{zhao2025urbanvideobench} probes embodied reasoning from egocentric flight video, but even its action items are multiple choice.
UAV-Track VLA, CognitiveDrone and ActiveFly-Bench learn the action mapping in the weights. We specify it in the prompt, so a failure may also be a failure to follow the specification, which Section~\ref{sec:experiments:gap} finds dominant below frontier scale.
ActiveFly-Bench chains question answering, observation planning and fine-grained control on a real quadrotor, but its ``hierarchy'' means levels of task abstraction inside one drone rather than an organisation of agents.
Real-drone results are reported quantitatively only by Fly0 and SPF, neither of which compares models in the air. OnFly, ActiveFly-Bench and the PX4 stack show flight demonstrations without reported metrics, and the remaining benchmarks are evaluated in simulation or on static data.
No prior benchmark in the table evaluates closed-loop control of several drones under one agent, and none compares several MLLMs on a real drone.

\paragraph{MissionBench-style agent on our benchmark.}
Table~\ref{tab:mbagent} runs the agent design of MissionBench with Gemini 3.7 Flash on the residential map, ten episodes per task type, one run each.
The agent keeps its eight relative-motion primitives and its \texttt{navigation done} action, its system prompt verbatim, and its per-step prompt with the pose, the last three actions, the three most recent frames and a bounding-box prediction.
Only the camera line, the action budget, the step and turn caps and the image references were adapted, and the simulator, episodes, judge and time limit are the same for both agents.
The two agents tie on approach, where the target is ahead and static, and separate on the tasks where the target is out of view or moving.
On the two moving-target tasks the MissionBench-style agent enters $\delta$ in three of twenty episodes and succeeds in none.

\begin{table}[t]
\centering
\caption{Coverage of prior work and \bench{} across seven axes. The axes are closed-loop control evaluation, moving-target tracking, out-of-view search, one agent driving $N$ drones, comparison across multiple MLLMs, analysis across model scales, and evaluation on a real drone.
\yes{} covered, \parti{} partially covered, \no{} not covered.
Out-of-view search is \yes{} when finding a target outside the field of view is an evaluated task. Robustness to losing sight of a target mid-episode is \parti{}. Model scale is \yes{} when one family is compared at three or more sizes. Two sizes are \parti{}. Real drone is \yes{} when results on a physical drone are reported with episode counts and a metric. A flight demonstration without reported metrics is \parti{}.
A further axis, commander + sub-agents (per-drone agents under a commander), is covered by no prior work, and \bench{} evaluates it in Section~\ref{sec:experiments:org}.
A mark is \yes{} only when the axis is a reported evaluation rather than a demonstration.}
\label{tab:coverage}
\input{tables/table_coverage.tex}
\end{table}

\begin{table}[!htb]
\centering
\caption{The agent design of MissionBench and ours on the \emph{single-drone} suite of the residential map, successes out of ten episodes per task type, Gemini 3.7 Flash, one run.}
\label{tab:mbagent}
\input{tables/table_mb_compare.tex}
\end{table}

\paragraph{Multi-drone agents.}
\label{app:relatedmulti}
Work beyond a single drone is sparse. AeroDuo~\citep{wu2025aeroduo} introduces a dual-altitude cooperative VLN task, and surveys list multi-UAV aerial VLN as open~\citep{chen2026uavvlnsurvey}.
AirCopBench~\citep{zha2026aircopbench} evaluates forty MLLMs on 14{,}610 multiple-choice items built from synchronised multi-view images, but its ``multi-drone collaborative'' setting is about understanding several views. No model in it issues an action. OnFly's ``dual-agent'' design likewise names two reasoning streams inside one drone, not two drones~\citep{zheng2026onfly}. RALLY~\citep{wang2025rally} does run a closed loop over multiple UAVs, but through one LLM per drone reasoning over structured local state, with separate control policies flying the platforms.
To our knowledge no prior work evaluates an MLLM controlling several drones from multi-view observations in a closed loop.

%% file: tables/table_coverage.tex
\providecommand{\covtabfont}{\footnotesize}
\providecommand{\covtabsep}{2.0pt}
\setlength{\tabcolsep}{\covtabsep}
\renewcommand{\arraystretch}{1.16}
\covtabfont
\begin{tabular}{@{}l ccc cccc@{}}
\toprule
& \multicolumn{3}{c}{\textbf{Task coverage}}
& \multicolumn{4}{c}{\textbf{Evaluation scope}} \\
\cmidrule(lr){2-4}\cmidrule(lr){5-8}
& Closed & Moving & Out-of-view & Multi-drone & Multiple & Model & Real \\
& loop   & target & search      & control  & MLLMs    & scale & drone \\
\midrule
OnFly~\citep{zheng2026onfly} & \yes  & \parti & \no   & \no   & \no   & \yes  & \parti  \\
SPF~\citep{hu2025spf} & \yes  & \yes  & \yes  & \no   & \yes  & \parti & \yes  \\
BEDI~\citep{guo2026bedi} & \yes  & \yes  & \no   & \no   & \yes  & \no  & \no  \\
DeTrack~\citep{hu2026detrack} & \yes  & \yes  & \parti & \no  & \no   & \no  & \no  \\
UAV-Track VLA~\citep{zhang2026uavtrackvla} & \yes  & \yes  & \no   & \no   & \no   & \no  & \no  \\
ActiveFly-Bench~\citep{zhang2026activefly} & \yes  & \no   & \yes  & \no   & \yes  & \no  & \parti  \\
AirCopBench~\citep{zha2026aircopbench} & \no   & \no   & \no   & \parti & \yes  & \parti  & \no  \\
UrbanVideo-Bench~\citep{zhao2025urbanvideobench} & \no   & \no   & \no   & \no   & \yes  & \yes  & \no  \\
PX4 Dialogue~\citep{lim2025px4dialogue} & \yes  & \no   & \yes  & \no   & \yes  & \no  & \parti  \\
UAV-ON~\citep{xiao2025uavon} & \yes  & \no   & \yes  & \no   & \no   & \no  & \no  \\
Fly0~\citep{xu2026fly0} & \yes  & \no   & \parti & \no  & \yes  & \parti  & \yes  \\
MissionBench~\citep{navaratnarajah2026zero} & \yes  & \no   & \no & \no  & \yes  & \yes  & \no  \\
\midrule
\rowcolor{cvGreen!8}
\textbf{DroneCATS (ours)}
                 & \yes  & \yes  & \yes  & \yes  & \yes & \yes  & \yes  \\
\bottomrule
\end{tabular}

%% file: tables/table_mb_compare.tex
\setlength{\tabcolsep}{6pt}
\renewcommand{\arraystretch}{1.12}
\footnotesize
\begin{tabular}{@{}lcc@{}}
\toprule
Task type & MissionBench-style agent & Ours (point-and-depth) \\
\midrule
Approach       & 9/10 & 9/10 \\
Search         & 4/10 & 7/10 \\
Track          & 0/10 & 5/10 \\
Search + track & 0/10 & 1/10 \\
\midrule
Total          & 13/40 & 22/40 \\
\bottomrule
\end{tabular}